\documentclass[sigconf, screen]{acmart}

\usepackage{tikz}
\usepackage{multirow}
\usepackage{subcaption}

\usepackage{amsmath,amsfonts,bm}

\def\figref#1{figure~\ref{#1}}
\def\Figref#1{Figure~\ref{#1}}

\def\secref#1{section~\ref{#1}}

\def\eqref#1{equation~(\ref{#1})}
\def\Eqref#1{Equation~(\ref{#1})}
\def\plaineqref#1{(\ref{#1})}
\def\tabref#1{table~\ref{#1}}
\def\Tabref#1{Table~\ref{#1}}

\def\1{\bm{1}}

\def\erva{{\textnormal{a}}}

\def\vc{{\bm{c}}}

\def\vg{{\bm{g}}}

\def\vs{{\bm{s}}}

\def\vv{{\bm{v}}}

\def\vz{{\bm{z}}}

\def\mA{{\bm{A}}}

\def\mW{{\bm{W}}}
\def\mX{{\bm{X}}}

\def\mZ{{\bm{Z}}}

\DeclareMathAlphabet{\mathsfit}{\encodingdefault}{\sfdefault}{m}{sl}
\SetMathAlphabet{\mathsfit}{bold}{\encodingdefault}{\sfdefault}{bx}{n}

\def\gA{{\mathcal{A}}}

\def\gD{{\mathcal{D}}}
\def\gE{{\mathcal{E}}}

\def\gG{{\mathcal{G}}}

\def\gI{{\mathcal{I}}}

\def\gL{{\mathcal{L}}}
\def\gM{{\mathcal{M}}}
\def\gN{{\mathcal{N}}}

\def\gR{{\mathcal{R}}}

\def\gV{{\mathcal{V}}}

\def\sR{{\mathbb{R}}}

\newcommand{\sigmoid}{\sigma}

\AtBeginDocument{%
  \providecommand\BibTeX{{%
    \normalfont B\kern-0.5em{\scshape i\kern-0.25em b}\kern-0.8em\TeX}}}

\setcopyright{acmcopyright}
\copyrightyear{2021}
\acmYear{2021}

\acmConference[CIKM '21]{30th ACM International Conference on Information and Knowledge Management}{1-5 November 2021}{Online}
\acmBooktitle{30th ACM International Conference on Information and Knowledge Management, 1-5 November 2021, Online}

\newcommand{\ours}{\textsc{VaCA-HINE} }
\begin{document}
\title[\ours]{A Framework for Joint Unsupervised Learning of Cluster-Aware Embedding for Heterogeneous Networks}

\author{Rayyan Ahmad Khan}
\affiliation{%
  \institution{Technical University of Munich}
  \city{Munich}
  \country{Germany}}
  \email{rayyan.khan@tum.de}
  \affiliation{%
  \institution{Mercateo AG}
  \city{Munich}
  \country{Germany}}
\email{RayyanAhmad.Khan@mercateo.com}

\author{Martin Kleinsteuber}
\affiliation{%
  \institution{Mercateo AG}
  \city{Munich}
  \country{Germany}}
  \email{kleinsteuber@tum.de}
  \affiliation{%
  \institution{Technical University of Munich}
  \city{Munich}
  \country{Germany}}
\email{Martin.Kleinsteuber@mercateo.com}

\renewcommand{\shortauthors}{Rayyan, et al.}


\begin{abstract}
  \footnote{Draft submitted to CIKM 2021}Heterogeneous Information Network (HIN) embedding refers to the low-dimensional projections of the HIN nodes that preserve the HIN structure and semantics.
  HIN embedding has emerged as a promising research field for network analysis as it enables downstream tasks such as clustering and node classification.
  In this work, we propose \ours for joint learning of cluster embeddings as well as cluster-aware HIN embedding.
  We assume that the connected nodes are highly likely to fall in the same cluster, and adopt a variational approach to preserve the information in the pairwise relations in a cluster-aware manner.
  In addition, we deploy contrastive modules to simultaneously utilize the information in multiple meta-paths, thereby alleviating the meta-path selection problem - a challenge faced by many of the famous HIN embedding approaches.
  The HIN embedding, thus learned, not only improves the clustering performance but also preserves pairwise proximity as well as the high-order HIN structure.
  We show the effectiveness of our approach by comparing it with many competitive baselines on three real-world datasets on clustering and downstream node classification.
\end{abstract}



\maketitle

\section{Introduction}
Network structures or graphs are used to model the complex relations between entities in a variety of domains such as biological networks\cite{applications-bio1,applications-bio2,applications-bio3}, social networks\cite{applications-social1,applications-social2,applications-social3,applications-social4,applications-social5}, chemistry\cite{applications-chem1,applications-chem2}, knowledge graphs\cite{applications-knowledgeGraphs1,applications-knowledgeGraphs2}, and many others\cite{applications-survey1,applications-survey2,applications-others1}.

\begin{figure}
    \input{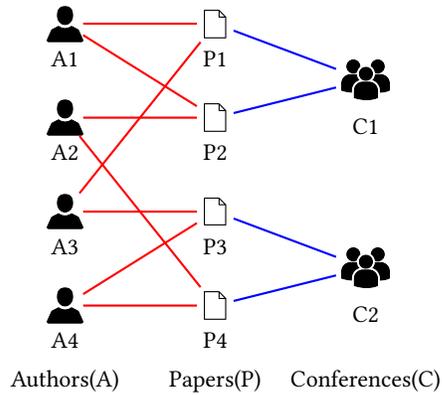}
    \caption{An example of HIN with three types of nodes and two types of edges.}
    \label{fig:hinExample}
\end{figure}

A large number of real-world phenomena express themselves in the form of heterogeneous information networks or HINs consisting of multiple node types and edges\cite{HINDef,metapathDef}.
For instance, \figref{fig:hinExample} illustrates a sample HIN consisting of three types of nodes (authors, papers, and conferences) and two types of edges.
A red edge indicates that an author \textit{has written} a paper and a blue edge models the relation that a paper \textit{is published in} a conference.
Compared to homogeneous networks (i.e., the networks consisting of only a single node type and edge type), HINs are able to convey a more comprehensive view of data by explicitly modeling the rich semantics and complex relations between multiple node types.
The advantages of HINs over homogeneous networks have resulted in an increasing interest in the techniques related to HIN embedding\cite{heterogeneous-embb-survey}.
The main idea of a network embedding is to project the graph nodes into a continuous low-dimensional vector space such that the structural properties of the network are preserved\cite{embeddings-survey1,embeddings-survey2}.
Many HIN embedding methods follow \textit{meta-paths} based approaches to preserve HIN structure\cite{related:metapath2vec,related:Hin2vec,related:heRec}, where a meta-path is a template indicating the sequence of relations between two node types in HIN.
As an example, PAP and PCP in \figref{fig:hinExample} are two meta-paths defining different semantic relations between two paper nodes.
Although the HIN embedding techniques based on meta-paths have received considerable success, meta-path selection is still an open problem.
Above that, the quality of HIN embedding highly depends on the selected meta-path\cite{related-just}.
To overcome this issue, either domain knowledge is utilized, which can be subjective and expensive, or some strategy is devised to fuse the information from predefined meta-paths\cite{related:Hin2vec,related-meta-path-fusion1,related-meta-path-fusion2}.

Node clustering is another important task for unsupervised network analysis, where the aim is to group similar graph nodes together.
The most basic approach to achieve this can be to apply an off-the-shelf clustering algorithm, e.g., K-Means or Gaussian Mixture Model (GMM), on the learned network embedding.
However, this usually results in suboptimal results because of treating the two highly correlated tasks, i.e., network embedding and node clustering, in an independent way.
There has been substantial evidence from the domain of euclidean data that the joint learning of latent embeddings and clustering assignments yields better results compared to when these two tasks are treated independently\cite{euclidean-joint1,euclidean-joint2,euclidean-joint3,euclidean-joint4,euclidean-joint5}.
For homogeneous networks, there are some approaches, e.g., \cite{gemsec,cnrl,jia2019communitygan}, that take motivation from the euclidean domain to jointly learn the network embedding and the cluster assignments.
However, no such approach exists for HINs to the best of our knowledge.

We propose an approach for \textbf{Va}riational \textbf{C}luster \textbf{A}ware \textbf{HIN} \textbf{E}mbedding, called \ours to address the above challenges of meta-path selection and joint learning of HIN embedding and cluster assignments.
Our approach makes use of two parts that simultaneously refine the target network embedding by optimizing their respective objectives.
The first part employs a variational approach to preserve pairwise proximity in a cluster-aware manner by aiming that the connected nodes fall in the same cluster and vice versa.
The second part utilizes a contrastive approach to preserve high-order HIN semantics by discriminating between real and corrupted instances of different meta-paths.
\ours is flexible in the sense that multiple meta-paths can be simultaneously used and they all aim to refine a single embedding.
Hence, the HIN embedding is learned by jointly leveraging the information in pairwise relations, the cluster assignments, and the high-order HIN structure.

Our major contributions are summarized below:
\begin{itemize}
    \item To the best of our knowledge, our work is the first to propose a unified approach for learning the HIN embedding and the node clustering assignments in a joint fashion.
    \item We propose a novel architecture that fuses together variational and contrastive approaches to preserve pairwise proximity as well as high order HIN semantics by simultaneously employing multiple meta-paths, such that the meta-path selection problem is also mitigated.
    \item We show the efficacy of our approach by conducting clustering and downstream node classification experiments on multiple datasets.
\end{itemize}
\section{Related Work}
\subsection{Unsupervised Network Embedding}
Most of the earlier work on network embedding targets homogeneous networks and employs random-walk based objectives, e.g., DeepWalk\cite{related:deepwalk}, Node2Vec\cite{related:node2vec}, and LINE\cite{related:line}.
Afterwards, the success of different graph neural network (GNN) architectures (e.g., graph convolutional network or GCN\cite{exp:split-2}, graph attention network or GAT\cite{gat} and GraphSAGE\cite{graph-sage}, etc.) gave rise to GNN based network embedding approaches\cite{embeddings-survey1,embeddings-survey2}.
For instance, graph autoencoders(GAE and VGAE\cite{vgae}) extend the idea of variational autoencoders\cite{vae-kingma} to graph datasets while deploying GCN modules to encode latent node embeddings as gaussian random variables.
Deep Graph Infomax (DGI \cite{dgi}) is another interesting approach that employs GCN encoder.
DGI extends the idea of Deep Infomax\cite{deep-infomax} to graph domain and learns network embedding in a contrastive fashion by maximizing mutual information between a graph-level representation and high-level node representations.
Nonetheless, these methods are restricted to homogeneous graphs and fail to efficiently learn the semantic relations in HINs.

Many HIN embedding methods, for instance Metapath2Vec\cite{related:metapath2vec}, HIN2Vec\cite{related:Hin2vec}, RHINE\cite{related-rhine} and HERec\cite{related:heRec}, etc., are partially inspired by homogeneous network embedding techniques in the sense that they employ meta-paths based random walks for learning HIN embedding.
In doing so, they inherit the known challenges of random walks such as high dependence on hyperparameters and sacrificing structural information to preserve proximity information \cite{ribeiro2017struc2vec}.
Moreover, their performance highly depends upon the selected meta-path or the strategy adopted to fuse the information from different meta-paths.
Recent literature also proposes HIN embedding using the approaches that do not depend on one meta-path.
For instance, a jump-and-stay strategy is proposed in \cite{related-just} for learning the HIN embedding.
HeGAN\cite{related:hegan} employs a generative adversarial approach for HIN embedding.
Following DGI\cite{dgi}, HDGI\cite{related-hetero-dgi} adopts a contrastive approach based on mutual information maximization.
NSHE\cite{nshe} is another approach to learn HIN embedding based on multiple meta-paths.
It uses two sets of embeddings; the node embeddings preserve edge-information and the context embeddings preserve network schema.
DHNE\cite{related:dhne} is a hyper-network embedding based approach that can be used to learn HIN embedding by considering meta-paths instances as hyper-edges.

\subsection{Node Clustering}
For homogeneous networks, we classify the approaches for node clustering in the following three classes:
\begin{enumerate}
    \item The most basic approach is the unsupervised learning of the network embedding, followed by a clustering algorithm, e.g., K-Means or GMM, on the embedding.
    \item Some architectures learn the network embedding with the primary objective to find good clustering assignments\cite{related-homo-cls1,related-homo-cls2}.
    However, they usually do not perform well on downstream tasks such as node classification as the embeddings are primarily aimed to find clusters.
    \item There are some other approaches, e.g., CommunityGAN\cite{jia2019communitygan}, CNRL\cite{cnrl} and GEMSEC\cite{gemsec}, etc., that take motivation from euclidean domain to jointly learn the network embedding and the cluster assignments.
\end{enumerate}
To the best of our knowledge, there exists no approach for HINs that can be classified as $(2)$ or $(3)$ as per the above list.
For HINs, all the known methods use the most basic approach, stated in $(1)$, to infer the cluster assignments from the HIN embedding.
It is worth noticing that devising an approach of class $(2)$ or $(3)$ for HINs is non-trivial as it requires explicit modeling of heterogeneity as well as a revision of the sampling techniques.
\ours is classified as $(3)$ as it presents a unified approach for cluster-aware HIN embedding that can also be utilized for downstream tasks such as node classification.

\section{Preliminaries}
This section sets up the basic definitions and notations that will be followed in the subsequent sections.

\begin{definition}[\textbf{Heterogeneous Information Network \cite{HINDef}}]
    A Heterogeneous Information Network or \textit{HIN} is defined as a graph $\gG = \{\gV, \gE, \gA, \gR, \pi, \lambda \}$ with the sets of nodes and edges represented by $\gV$ and $\gE$ respectively. 
    $\gA$ and $\gR$ denote the sets of node-types and edge-types respectively. 
    In addition, a HIN has a node-type mapping function $\pi : \gV \rightarrow \gA$ and an edge-type mapping function $\lambda : \gE \rightarrow \gR$ such that $|\gA| + |\gR| > 2$.
\end{definition}

\textbf{Example: } \Figref{fig:hinExample} illustrates a sample HIN on an academic network, consisting of three types of nodes, i.e., authors (A), papers (P), and conferences (C).
Assuming symmetric relations between nodes, we end up with two types of edges, i.e., the author-paper edges, colored in red, and the paper-conference edges, colored in blue.

\begin{definition}[\textbf{Meta-path \cite{metapathDef}}]
    A meta-path $\gM$ of length $|\gM|$ is defined on the graph network schema as a sequences of the form
    $$A_1 \xrightarrow[]{R_1} A_2 \xrightarrow[]{R_2} \cdots \xrightarrow[]{R_{|\gM|}} A_{|\gM| + 1},$$
    where $A_i \in \gA$ and $R_i \in \gR$.
    Using $\circ$ to denote the composition operator on relations, a meta-path can be viewed as a composite relation $R_1 \circ R_2 \cdots \circ R_{|\gM|}$ between $A_1$ and $A_{|\gM| + 1}$.
    A meta-path is often abbreviated by the sequence of the involved node-types i.e. ($A_1 A_2 \cdots A_{|\gM| + 1}$).
\end{definition}

\textbf{Example: } In \figref{fig:hinExample}, although no direct relation exists between authors, we can build semantic relations between them based upon the co-authored papers. 
This results in the meta-path (APA). 
We can also go a step further and define a longer meta-path (APCPA) by including conference nodes.
Similarly, semantics between two papers can be defined in terms of the same authors or same conferences, yielding two possible meta-paths, i.e., (PAP) and (PCP).

In this work, we denote the $p$-th meta-path by $\gM_p$ and the set of the samples of $\gM_p$ is denoted by $\{m_p\}$ where $m_p$ refers to a single sample of $\gM_p$. The node at $n$-th position in $m_p$ is denoted by $m_p(n)$.

\section{Problem Formulation}
Suppose a HIN $\gG = \{\gV, \gE, \gA, \gR, \pi, \lambda \}$ and a matrix $\mX \in \sR^{N \times F}$ of $F$-dimensional node features, $N$ being the number of nodes.
Let $\mA \in \{0, 1\}^{N \times N}$ denote the adjacency matrix of $\gG$, with $\erva_{ij}$ referring to the element in $i$-th row and $j$-th column of $\mA$.
Given $K$ as the number of clusters, our aim is to jointly learn the $d$-dimensional cluster embeddings as well as the HIN embedding such that these embeddings can be used for clustering as well as downstream node classification.

\begin{figure*}
    \includegraphics[width=\linewidth]{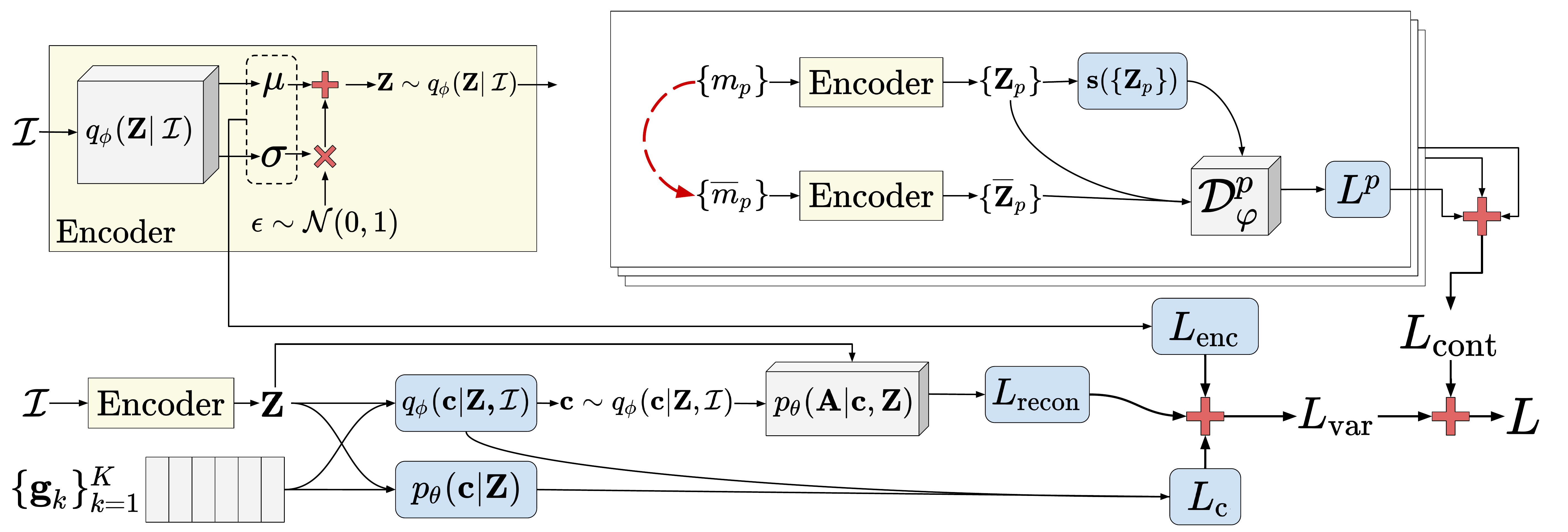}
    \caption{
        Overview of \ours architecture.
        For illustrational convenience, the variational encoder block has been extracted to the top-left.
        All the encoder blocks have been colored yellow to highlight that a single shared encoder is used in the whole architecture.
        Blocks containing learnable and non-learnable parameters are colored gray and blue respectively.
        On the bottom, lies the variational module for learning the node embeddings $\mZ$ and the cluster embeddings $\{\vg_k\}_{k = 1}^K$ such that the variational loss $L_{\mathrm{var}}$, given in \eqref{eq:LvarTerms}, is minimized.
        The $p$-th contrastive module lies on the top right.
        It discriminates between true/positive and corrupted/negative samples for the meta-path $\gM_p$.
        The red-dashed arrow indicates the corruption of the positive samples $\{m_p\}$ of the meta-path $\gM_p$, to generate negative samples $\{\overline{m}_p\}$.
        The encoded representations of the true and corrupted versions of the respective meta-path samples are denoted by $\{\mZ_p\}$ and $\{\overline{\mZ}_p\}$.
        These representations, along with the summary vectors $\vs\big(\{\mZ_p\}\big)$, are fed to the discriminator $\gD_{\varphi}^p$ to distinguish between positive and negative samples.
    }
    \label{fig:modelOverview}
\end{figure*}

\Figref{fig:modelOverview} gives an overview of the \ours framework.
It consists of a variational module and a set of contrastive modules.
The main goal of the variational module is to optimize the reconstruction of $\mA$ by simultaneously incorporating the information in the HIN edges and the clusters.
The contrastive modules aim to preserve the high-order structure by discriminating between the positive and negative samples of different meta-paths.
So, if $M$ meta-paths are selected to preserve high-order HIN structure, the overall loss function, to be minimized, takes the following form

\begin{align}
    L & = L_{\mathrm{var}} + L_{\mathrm{cont}} \label{eq:abstractLoss}                  \\
      & = L_{\mathrm{var}} + \sum \limits_{p = 1}^M L^p \label{eq:abstractLossWithSum}
\end{align}
where $L_{\mathrm{var}}$ and $L_{\mathrm{cont}}$ refer to the loss of variational module and contrastive modules respectively. We now discuss these modules along with their losses in detail.

\section{Variational Module}\label{sec:variationalModule}
The objective of the variational module is to recover the adjacency matrix $\mA$.
More precisely, we aim to learn the free parameters $\theta$ of our model such that $\mathrm{log}\big(p_{\theta}(\mA)\big)$ is maximized.

\subsection{Generative Model}
Let the random variables $\vz_i$ and $c_i$ respectively denote the latent node embeddings and the cluster assignment for the $i$-th node.
The generative model is then given by
\begin{align}
    p_{\theta}(\mA) & = \int \sum \limits_{\vc} p_{\theta}(\mZ, \vc, \mA) d\mZ, \label{eq:generativeModelJoint}
\end{align}
where
\begin{align}
    \vc & = [c_1, c_2, \cdots, c_N],       \\
    \mZ & = [\vz_1, \vz_2, \cdots, \vz_N].
\end{align}
So, $\vc$ and $\mZ$ stack cluster assignments and node embeddings respectively.
We factorize the joint distribution in \eqref{eq:generativeModelJoint} as
\begin{equation}
    p_{\theta}(\mZ, \vc, \mA) = p(\mZ) p_{\theta}(\vc | \mZ) p_{\theta}(\mA | \vc, \mZ). \label{eq:generativeModelFactorizedJoint}
\end{equation}
In order to further factorize \eqref{eq:generativeModelFactorizedJoint}, we make the following assumptions:
\begin{itemize}
    \item Following the existing approaches, e.g., \citep{vgae,khan2020epitomic}, we consider $\vz_i$ to be \textit{i.i.d.} random variables.
    \item The conditional random variables $c_i | \vz_i$ are also assumed \textit{i.i.d}.
    \item The reconstruction of $\erva_{ij}$ depends upon the node embeddings $\vz_i$ and $\vz_j$, as well as the cluster assignments $c_i$ and $c_j$.
          The underlying intuition comes from the observation that the connected nodes have a high probability of falling in the same cluster.
          Since $\vz_i$ have been assumed \textit{i.i.d.}, we need to reconstruct $\erva_{ij}$ in a way to ensure the dependence between the connected nodes and the clusters chosen by them.
\end{itemize}
Following the above assumptions, the distributions in \eqref{eq:generativeModelFactorizedJoint} can now be further factorized as
\begin{align}
    p(\mZ)                     & = \prod \limits_{i = 1}^{N} p(\vz_{i}), \label{eq:joint-Z}                                                 \\
    p_{\theta}(\vc | \mZ)      & = \prod \limits_{i = 1}^{N} p_{\theta}(c_{i} | \vz_{i}), \label{eq:joint-cGivenZ}                          \\
    p_{\theta}(\mA | \vc, \mZ) & =  \prod \limits_{i, j} p_{\theta}(\erva_{ij} | c_{i}, c_{j}, \vz_{i}, \vz_{j}). \label{eq:jointDecoder}
\end{align}

\subsection{Inference Model}
For notational convenience, we define $\gI = (\mA, \pi, \lambda, \mX)$.
To ensure computational tractability in \eqref{eq:generativeModelJoint}, we introduce the approximate posterior
\begin{align}
    q_{\phi}(\mZ, \vc | \gI) & = \prod \limits_i q_{\phi}(\vz_{i}, c_{i} | \gI)                                                      \\
                             & = \prod \limits_i q_{\phi} (\vz_{i} | \gI) q_{\phi} (c_{i} | \vz_{i}, \gI), \label{eq:jointPosterior}
\end{align}
where $\phi$ denotes the parameters of the inference model.
The objective now takes the form
\begin{align}
	\text{log}(p_{\theta}(\mA)) &= \text{log}\bigg( \mathbb{E}_{q_{\phi}(\mZ, \vc | \gI)} \bigg\{\frac{p(\mZ) \ p_{\theta}(\vc | \mZ) \ p_{\theta}(\mA | \vc, \mZ)}{q_{\phi}(\mZ | \gI) q_{\phi}(\vc | \mZ, \gI)}\bigg\} \bigg) \\
	&\geq \mathbb{E}_{q_{\phi}(\mZ, \vc | \gI)} \bigg\{ \text{log}\bigg( \frac{p(\mZ) \ p_{\theta}(\vc | \mZ) \ p_{\theta}(\mA | \vc, \mZ)}{q_{\phi}(\mZ | \gI) q_{\phi}(\vc | \mZ, \gI)}\bigg)\bigg\} \label{eq:obj:jensenApplied} \\
    &= \gL_{\mathrm{ELBO}},
\end{align}
where \plaineqref{eq:obj:jensenApplied} follows from Jensen's Inequality.
So we maximize, with respect to the parameters $\theta$ and $\phi$, the corresponding ELBO bound, given by
\begin{align}
    \gL_{\mathrm{ELBO}} & \approx  - \underbrace{\sum \limits_{i=1}^{N} D_{KL} (q_{\phi}(\vz_{i} | \gI) \ || \ p(\vz_{i}))}_{L_{\mathrm{enc}}}\nonumber                                                                                                                   \\
    -                      & \underbrace{\sum \limits_{i=1}^{N} \frac{1}{R} \sum \limits_{r = 1}^{R} D_{KL} (q_{\phi}(c_{i} | \vz_{i}^{(r)}, \gI) \ ||\ p_{\theta}(c_{i} | \vz_{i}^{(r)}))}_{L_{\mathrm{c}}} \nonumber                                        \\
    +                      & \underbrace{\sum \limits_{(i, j) \in \gE} \mathbb{E}_{q_{\phi}(\vz_{i}, \vz_{j}, c_{i}, c_{j} | \gI)}\bigg\{\text{log}\bigg(p_{\theta}(\erva_{ij} | c_{i}, c_{j}, \vz_{i}, \vz_{j})\bigg)\bigg\}}_{- L_{\mathrm{recon}}}, \label{eq:elbo}
\end{align}
where $D_{KL}(\cdot ||\cdot)$ represents the KL-divergence between two distributions.
The detailed derivation of \eqref{eq:elbo} is provided in the supplementary material.
\Eqref{eq:elbo} contains three major summation terms as indicated by the braces under these terms.
\begin{enumerate}
    \item The first term, denoted by $L_{\mathrm{enc}}$, refers to the encoder loss.
    Minimizing this term minimizes the mismatch between the approximate posterior and the prior distributions of the node-embeddings.
    \item The second term, denoted by $L_{\mathrm{c}}$, gives the mismatch between the categorical distributions governing the cluster assignments.
    Minimizing this term ensures that the cluster assignments $c_i$ take into consideration not only the respective node embeddings $\vz_i$ but also the high order HIN structure as given in $\gI$.
    \item The third term is the negative of the reconstruction loss or $L_{\mathrm{recon}}$.
    It can be viewed as the negative of binary cross-entropy (BCE) between the input and the reconstructed edges.
\end{enumerate}

Instead of maximizing the ELBO bound, we can minimize the corresponding loss, which we refer to as the variational loss or $L_{\mathrm{var}}$, given by
\begin{align}
    L_{\mathrm{var}} & = - \gL_{\mathrm{ELBO}} \\
                     & = L_{\mathrm{enc}} + L_{\mathrm{c}} + L_{\mathrm{recon}} \label{eq:LvarTerms}
\end{align}

\subsection{Choice of Distributions}\label{sec:distChoices}
Now we go through the individual loss terms of $L_{\mathrm{var}}$ in \eqref{eq:LvarTerms} and \plaineqref{eq:elbo}, and describe our choices of the underlying distributions.

\subsubsection{\textbf{Distributions Involved in} $L_{\mathrm{enc}}$}
The prior $p(\vz_i)$ in \eqref{eq:joint-Z} is fixed as the standard gaussian.
The respective approximate posterior $q_{\phi}(\vz_{i} | \gI)$ in \eqref{eq:jointPosterior} is also chosen to be a gaussian, with the parameters (mean $\bm{\mu}_{i}$ and variance $\bm{\sigma}^2_{i}$) learned by the encoder block. i.e.
\begin{align}
	q_{\phi} (\vz_{i} | \gI) = \gN\big(\bm{\mu}_{i}(\gI), \text{diag}({\bm{\sigma}^2}_{i}(\gI))\big). \label{eq:qzGivenI}
\end{align}

\subsubsection{\textbf{Distributions Involved in} $L_{\mathrm{c}}$}
For $K$ as the desired number of clusters, we introduce the cluster embeddings $\{\vg_1, \vg_2, \cdots, \vg_K\}$, $\vg_k \in \sR^d$. The prior $p_{\theta}(c_i | \vz_i)$ is then parameterized in the form of a softmax of the dot-products between node embeddings and cluster embeddings as
\begin{align}
	p_{\theta}(c_{i} & = k| \vz_{i}) = \mathrm{softmax}(\vz_{i}^T\vg_{k}). \label{eq:pcGivenZ}
\end{align}
The softmax is over $K$ cluster embeddings to ensure that $p_{\theta}(c_{i} | \vz_{i})$ is a distribution.

The corresponding approximate posterior $q_{\phi} (c_{i} | \vz_{i}, \gI)$ in \eqref{eq:jointPosterior} is conditioned on the node embedding $\vz_{i}$ as well as the high-order HIN structure.
The intuition governing its design is that if $i$-th node falls in $k$-th cluster, its immediate neighbors as well as high order neighbors have a relatively higher probability of falling in the $k$-th cluster, compared to the other nodes.
To model this mathematically, we make use of the samples of the meta-paths involving the $i$-th node.
Let us denote this set as $\zeta_i = \{m \mid \exists n:m(n) = i\}$.
For every meta-path sample, we average the embeddings of the nodes constituting the sample.
Afterward, we get a single representative embedding by averaging over $\zeta_i$ as

\begin{equation}
    \hat{\vz}_{i} = \frac{1}{|\zeta_i|} \sum \limits_{m \in \zeta_i} \frac{1}{|m| + 1} \sum \limits_{n=1}^{|m| + 1} \vz_{m(n)}. \label{eq:zAvgMetaPaths}
\end{equation}
The posterior distribution $q_{\phi} (c_{i} | \vz_{i}, \gI)$ can now be modeled similar to \eqref{eq:pcGivenZ} as
\begin{align}
	q_{\phi}(c_{i} = k| \vz_{i}, \gI) = \mathrm{softmax}(\hat{\vz}_{i}^T\vg_{k}). \label{eq:qcGivenZI}
\end{align}
So, while \eqref{eq:pcGivenZ} targets only $\vz_i$ for $c_i$, the \eqref{eq:qcGivenZI} relies on the high order HIN structure by targeting meta-path based neighbors.
Minimizing the KL-divergence between these two distributions ultimately results in an agreement where nearby nodes tend to have the same cluster assignments and vice versa.

\subsubsection{\textbf{Distributions Involved in} $L_{\mathrm{recon}}$}
The posterior distribution $q_{\phi}(\vz_{i}, \vz_{j}, c_{i}, c_{j} | \gI)$ in the third summation term of \eqref{eq:elbo} is factorized into two conditionally independent distributions  i.e.
\begin{align}
    q_{\phi}(\vz_{i}, \vz_{j}, c_{i}, c_{j} | \gI) = q_{\phi}(\vz_{i}, c_{i} | \gI) q_{\phi}(\vz_{j}, c_{j} | \gI).
\end{align}
where $q_{\phi}(\vz_{i}, c_{i} | \gI)$ factorization and the related distributions have been given in \eqref{eq:jointPosterior}, \plaineqref{eq:qzGivenI} and \plaineqref{eq:qcGivenZI}.

Finally, the \textit{edge-decoder} in \eqref{eq:jointDecoder} is modeled to maximize the probability of connected nodes to have same cluster assignments, i.e.,
\begin{align}
	p_{\theta}(\erva_{ij} | c_{i} = \ell, c_{j} = m, \vz_{i}, \vz_{j}) = \frac{\sigmoid(\vz_{i}^T\vg_{m}) + \sigmoid(\vz_{j}^T\vg_{\ell})}{2}, \label{eq:decoderDist}
\end{align}
where $\sigmoid(\cdot)$ is the sigmoid function.
The primary objective of the variational module in \eqref{eq:LvarTerms} is not to directly optimize the cluster assignments, but to preserve edge information in a cluster-aware fashion.
So the information in the cluster embeddings, the node embeddings, and the cluster assignments is simultaneously incorporated by \eqref{eq:decoderDist}.
This formation forces the cluster embeddings of the connected nodes to be similar and vice versa.
On one hand, this helps in learning better node representations by leveraging the global information about the graph structure via cluster assignments.
On the other hand, this also refines the cluster embeddings by exploiting the local graph structure via node embeddings and edge information.

\subsection{Practical Considerations}
Following the same pattern as in \secref{sec:distChoices}, we go through the individual loss terms in $L_{\mathrm{var}}$ and discuss the practical aspects related to minimization of these terms.
\subsubsection{\textbf{Considerations Related to} $L_{\mathrm{enc}}$}\label{sec:LencConsiderations}
For computational stability, we learn $\mathrm{log}(\sigma)$ instead of variance in \plaineqref{eq:qzGivenI}.
In theory, the parameters of $q_{\phi}(\vz_i | \gI)$ can be learned by any suitable network, e.g., relational graph convolutional network (RGCN \cite{rgcn}), graph convolutional network (GCN \cite{exp:split-2}), graph attention network (GAT \cite{gat}), GraphSAGE \cite{graph-sage}, or even two linear modules.
In practice, some encoders may give better HIN embedding than others.
Supplementary material\cite{supplementary} contains further discussion on this.
For efficient back-propagation we use the reparameterization trick\cite{vae-tutorial} as illustrated in the encoder block in \figref{fig:modelOverview}.

\subsubsection{\textbf{Considerations Related to} $L_{\mathrm{c}}$}
Since the community assignments in \eqref{eq:qcGivenZI} follow a categorical distribution, we use Gumbel-softmax\cite{gumbel-softmax} for efficient back-propagation of the gradients.
In the special case where a node $i$ does not appear in any meta-path samples, we formulate $\hat{\vz}_i$ in \plaineqref{eq:zAvgMetaPaths} by averaging over its immediate neighbors. i.e.
\begin{align}
    \hat{\vz}_i = \frac{1}{|\{j: \erva_{ij} = 1\}|}\sum \limits_{j: \ \erva_{ij} = 1} \vz_j,
\end{align}
i.e., when meta-path based neighbors are not available for a node, we restrict ourselves to leveraging the information in first-order neighbors only.

\subsubsection{\textbf{Considerations Related to} $L_{\mathrm{recon}}$}
The decoder block requires both positive and negative edges for learning the respective distribution in \eqref{eq:decoderDist}.
Hence we follow the current approaches, e.g., \cite{vgae,khan2020epitomic} and sample an equal number of negative edges from $\mA$ to provide at the decoder input along with the positive edges.

\section{Contrastive Modules}\label{sec:contrastiveRegularizationModules}
In addition to the variational module, \ours architecture consists of $M$ contrastive modules, each referring to a meta-path.
All the contrastive modules target the same HIN embedding $\mZ$ generated by the variational module.
Every module contributes to the enrichment of the HIN embedding by aiming to preserve the high-order HIN structure corresponding to a certain meta-path.
This, along with the variational module, enables \ours to learn HIN embedding by exploiting the local information (in the form of pairwise relations) as well as the global information (in the form of cluster assignments and samples of different meta-paths).
In this section, we discuss the architecture and the loss of $p$-th module as shown in the \figref{fig:modelOverview}.

\subsection{Architecture of $p$-th Contrastive Module}

\subsubsection{Input}
The input to $p$-th module is the set $\{m_p\}$ consisting of the samples of the meta-path $\gM_p$.

\subsubsection{Corruption}
As indicated by the red-dashed arrow in \figref{fig:modelOverview}, a corruption function is used to generate the set $\{\overline{m}_p\}$ of negative/corrupted samples using the HIN information related to true samples, where $\overline{m}_p$ is a negative sample corresponding to $m_p$.
\ours performs corruption by permuting the matrix $\mX$ row-wise to shuffle the features between different nodes.

\subsubsection{Encoder}
All the contrastive blocks in \ours use the same encoder, as used in the variational module.
For a sample $m_p$, the output of encoder is denoted by the matrix $\mZ_p \in \sR^{(|\gM_p| + 1) \times d}$ given by
\begin{align}
    \mZ_p = [\vz_{m_p(1)}, \vz_{m_p(2)}, \cdots, \vz_{m_p(|\gM_p| + 1)}] \label{eq:ZpMatrixColumns}
\end{align}
Following the same approach, we can obtain $\overline{\mZ}_p$ for a corrupted sample $\overline{m}_p$.

\subsubsection{Summaries}
The set of the representations of $\{m_p\}$, denoted by $\{\mZ_p\}$, is used to generate the \textit{summary} vectors $\vs(m_p)$ given by
\begin{align}
    \vs(m_p) = \frac{1}{|\gM_p| + 1}\sum \limits_{n = 1}^{|\gM_p| + 1} \vz_{m_p(n)}. \label{eq:mpSummaryVector}
\end{align}

\subsubsection{Discriminator}
The discriminator of $p$-th block is denoted by $\gD^p_{\varphi}$ where $\varphi$ collectively refers to the parameters of all the contrastive modules.
The aim of $\gD^p_{\varphi}$ is to differentiate between positive and corrupted samples of $\gM_p$.
It takes the representations $\{\mZ_p\}$ and $\{\overline{\mZ}_p\}$ along with the summary vectors at the input, and outputs a binary decision for every sample, i.e., $1$ for positive and $0$ for negative or corrupted samples.
Specifically, for a sample $m_p$, the discriminator $\gD^p_\varphi$ first flattens $\mZ_p$ in \eqref{eq:ZpMatrixColumns} into $\vv_{p}$ by vertically stacking the columns of $\mZ_p$ as
\begin{align}
    \vv_p = \mathop{\Vert} \limits_{n=1}^{|\gM_p| + 1} \vz_{m_p(n)},
\end{align}
where $\Vert$ denotes the concatenation operator.
Afterwards the output of $\gD^p_\varphi$ is given by
\begin{align}
    \gD^p_{\varphi}\Big(m_p, \vs(m_p)\Big) = \sigmoid\Big(\vv_p^T \mW_p \vs(m_p)\Big), \label{eq:DpDecisionPos}
\end{align}
where $\mW_p \in \sR^{(|\gM_p| + 1) \times d}$ is the learnable weight matrix of $\gD^p_\varphi$.
Similarly, for negative sample $\overline{m}_p$,
\begin{align}
    \gD^p_{\varphi}\Big(\overline{m}_p, \vs(m_p)\Big) = \sigmoid\Big(\overline{\vv}_p^T \mW_p \vs(m_p)\Big), \label{eq:DpDecisionNeg}
\end{align}
where $\overline{\vv}_p^T$ is the flattened version of $\overline{\mZ}_p$.

\subsection{Loss of $p$-th Contrastive Module}
The outputs in \eqref{eq:DpDecisionPos} and \plaineqref{eq:DpDecisionNeg} can be respectively viewed as the probabilities of $m_p$ and $\overline{m}_p$ being positive samples according to the discriminator.
Consequently, we can model the contrastive loss $L^p$, associated with the $p$-th block, in terms of the binary cross-entropy loss between the positive and the corrupted samples.
Considering $J$ negative samples $\overline{m}_p^{(j)}$ for every positive sample $m_p$, the loss $L^p$ can be written as
\begin{align}
    L^p & = -\frac{1}{(J + 1) \times |\{m_p\}|} \sum \limits_{m_p \in \{m_p\}} \Bigg( \mathrm{log} \Big[\gD^p_{\varphi}\Big(m_p, \vs(m_p)\Big)\Big] \nonumber \\
        & + \sum \limits_{j = 1}^{J} \mathrm{log} \Big[1 - \gD^p_{\varphi}\Big(\overline{m}_p^{(j)}, \vs(m_p)\Big)\Big]\Bigg). \label{eq:LpWithJNegativeSamples}
\end{align}

Minimizing $L^p$ preserves the high-order HIN structure by ensuring that the correct samples of $\gM_p$ are distinguishable from the corrupted ones.
This module can be replicated for different meta-paths.
Hence, given $M$ meta-paths, we have $M$ contrastive modules, each one aiming to preserve the integrity of the samples of a specific meta-path by using a separate discriminator.
The resulting loss, denoted by $L_{\mathrm{cont}}$, is the sum of the losses from all the contrastive modules as defined in \eqref{eq:abstractLossWithSum}.
Since the encoder is shared between the variational module and the contrastive modules, minimizing $L_{\mathrm{cont}}$ ensures that the node embeddings also leverage the information in the high-order HIN structure.
\section{Experiments}
In this section, we conduct an extensive empirical evaluation of our approach.
We start with a brief description of the datasets and the baselines selected for evaluation.
Afterward, we provide the experimental setup for the baselines as well as for \ours.
Then we analyze the results on clustering and downstream node classification tasks.
Furthermore, the supplementary material\cite{supplementary} contains the evaluation with different encoder modules both with and without contrastive loss $L_{\mathrm{cont}}$.

\subsection{Datasets}
We select two academic networks and a subset of IMDB\cite{heteroGat} for our experiments.
The detailed statistics of these datasets are given in \tabref{tab:datasets}.

\begin{table*}[tbh]
    \resizebox{0.7\linewidth}{!}{%
        \begin{tabular}{c||c|c||c|c||c}
            \toprule
            \textbf{Dataset}               & \textbf{Nodes}       & \textbf{Number of Nodes} & \textbf{Relations}                         & \textbf{Number of Relations} & \textbf{Used Meta-paths} \\
            \midrule
            \multirow{3}{*}{DBLP} & Papers (P)      & 9556            & \multirow{3}{*}{\shortstack[]{P-A \\ P-C}} & \multirow{3}{*}{\shortstack[]{18304 \\ 9556}} & ACPCA        \\
                                  & Authors (A)     & 2000            & & & APA        \\
                                  & Conferences (C) & 20              &                                   &                     & ACA        \\ \hline
            \multirow{3}{*}{ACM}  & Papers (P)      & 4019            & \multirow{3}{*}{\shortstack[]{P-A \\ P-S}} & \multirow{3}{*}{\shortstack[]{13407 \\ 4019}} & \multirow{3}{*}{\shortstack[]{PAP \\ PSP}}        \\
                                  & Authors (A)     & 7167            &                                   &                     &         \\
                                  & Subjects (S)    & 60              &                                   &                     &            \\ \hline
            \multirow{3}{*}{IMDB} & Movies (M)      & 3676            & \multirow{3}{*}{\shortstack[]{M-A \\ M-D}} & \multirow{3}{*}{\shortstack[]{11028 \\ 3676}} & \multirow{3}{*}{\shortstack[]{MAM \\ MDM}}        \\
                                  & Actors (A)      & 4353            & & & \\
                                  & Directors (D)   & 1678            &                                   &                     &            \\
            \bottomrule
        \end{tabular}%
    }
    \caption{Statistics of the datasets used for evaluation.}
    \label{tab:datasets}
\end{table*}

\subsubsection{\textbf{DBLP\cite{rgcn}}}
The extracted subset of DBLP\footnote{\url{https://dblp.uni-trier.de/}} has the author and paper nodes divided into four areas, i.e., database, data mining, machine learning, and information retrieval.
We report the results of clustering and node classification for author nodes as well as paper nodes by comparing with the research area as the ground truth.
We do not report the results for conference nodes as they are only 20 in number.
To distinguish between the results of different node types, we use the names DBLP-A and DBLP-P respectively for the part of DBLP with the node-type authors and papers.

\subsubsection{\textbf{ACM\cite{heteroGat}}}
Following \cite{heteroGat} and \cite{nshe}, the extracted subset of ACM\footnote{\url{http://dl.acm.org/}} is chosen with papers published in KDD, SIGMOD, SIGCOMM, MobiCOMM and VLDB.
The paper features are the bag-of-words representation of keywords.
They are classified based upon whether they belong to data-mining, database, or wireless communication.

\subsubsection{\textbf{IMDB\cite{heteroGat}}}
Here we use the extracted subset of movies classified according to their genre, i.e., action, comedy, and drama.
The dataset details along with the meta-paths, used by \ours as well as other competitors, are given in \tabref{tab:datasets}.

\subsection{Baselines}
We compare the performance of \ours with $15$ competitive baselines used for unsupervised network embedding.
These baselines include $5$ homogeneous network embedding methods, $3$ techniques proposed for joint homogeneous network embedding and clustering, and $7$ HIN embedding approaches.

\subsubsection{\textbf{DeepWalk\cite{related:deepwalk}}}
It learns the node embeddings by first performing classical truncated random walks on an input graph, followed by the skip-gram model.

\subsubsection{\textbf{LINE\cite{related:line}}}
This approach samples node pairs directly from a homogeneous graph and learns node embeddings with the aim of preserving first-order or second-order proximity, respectively denoted by LINE-1 and LINE-2 in this work.

\subsubsection{\textbf{GAE\cite{vgae}}}
GAE extends the idea of autoencoders to graph datasets.
The aim here is to reconstruct $\mA$ for the input homogeneous graph.

\subsubsection{\textbf{VGAE\cite{vgae}}}
This is the variational counterpart of GAE.
It models the latent node embeddings as Gaussian random variables and aims to optimize $\mathrm{log}\big(p(\mA)\big)$ using a variational model.

\subsubsection{\textbf{DGI\cite{dgi}}}
Deep-Graph-Infomax extends Deep-Infomax\cite{deepinfomax} to graphs.
This is a contrastive approach to learn network embedding such that the true samples share a higher similarity to a global network representation (known as \textit{summary}), as compared to the corrupted samples.

\subsubsection{\textbf{GEMSEC\cite{gemsec}}}
This approach jointly learns network embedding and node clustering assignments by sequence sampling.

\subsubsection{\textbf{CNRL\cite{cnrl}}}
CNRL makes use of the node sequences generated by random-walk based techniques, e.g., DeepWalk, node2vec, etc., to jointly learn node communities and node embeddings.

\subsubsection{\textbf{CommunityGAN\cite{jia2019communitygan}}}
As the name suggests, it is a generative adversarial approach for learning community based network embedding.
Given $K$ as the number of desired communities, it learns $K$-dimensional latent node embeddings such that every dimension gives the assignment strength for the target node in a certain community.

\subsubsection{\textbf{Metapath2Vec\cite{related:metapath2vec}}}
This HIN embedding approach makes use of a meta-path when performing random walks on graphs.
The generated random walks are then fed to a heterogeneous skip-gram model, thus preserving the semantics-based similarities in a HIN.

\subsubsection{\textbf{HIN2Vec\cite{related:Hin2vec}}}
It jointly learns the embeddings of nodes as well as meta-paths, thus preserving the HIN semantics.
The node embeddings are learned such that they can predict the meta-paths connecting them.

\subsubsection{\textbf{HERec\cite{related:heRec}}}
It learns semantic-preserving HIN embedding by designing a type-constraint strategy for filtering the node-sequences based upon meta-paths.
Afterward, skip-gram model is applied to get the HIN embedding.

\subsubsection{\textbf{HDGI\cite{related-hetero-dgi}}}
HDGI extends DGI model to HINs.
The main idea is to disassemble a HIN to multiple homogeneous graphs based upon different meta-paths, followed by semantic-level attention to aggregate different node representations.
Afterward, a contrastive approach is applied to maximize the mutual information between the high-level node representations and the graph representation.

\subsubsection{\textbf{DHNE\cite{related:dhne}}}
This approach learns hyper-network embedding by modeling the relations between the nodes in terms of indecomposable hyper-edges.
It uses deep autoencoders and classifiers to realize a non-linear tuple-wise similarity function while preserving both local and global proximities in the formed embedding space.

\subsubsection{\textbf{NSHE\cite{nshe}}}
This approach embeds a HIN by jointly learning two embeddings: one for optimizing the pairwise proximity, as dictated by $\mA$, and the second one for preserving the high-order proximity, as dictated by the meta-path samples.

\subsubsection{\textbf{HeGAN\cite{related:hegan}}}
HeGAN is a generative adversarial approach for learning HIN embedding by discriminating between real and fake heterogeneous relations.

\subsection{Implementation Details}

\subsubsection{\textbf{For Baselines}}
The embedding dimension $d$ for all the methods is kept to $128$.
The only exception is CommunityGAN because it requires the node embeddings to have the same dimensions as the desired clusters, i.e., $d = K$.
The rest of the parameters for all the competitors are kept the same as given by their authors.
For GAE and VGAE, the number of sampled negative edges is the same as the positive edges as given in the original implementations.
For the competitors that are designed for homogeneous graphs (i.e., DeepWalk, LINE, GAE, VGAE, and DGI), we treat the HINs as homogeneous graphs.
For the methods Metapath2Vec and HERec, we test all the meta-paths in \tabref{tab:datasets} and report the best results.
For DHNE, the meta-path instances are considered as hyper-edges.
The embeddings obtained by DeepWalk are used for all the models that require node features.

\subsubsection{\textbf{For \ours}}
The training process of \ours has the following steps:
\begin{enumerate}
    \item \textbf{$\mZ$ Initialization: }
          This step involves pre-training of the variational encoder to get $\mZ$.
          So, $L_{\mathrm{enc}}$ is the only loss considered in this step.
    \item \textbf{$\vg_k$ Initialization: }
          We fit K-Means on $\mZ$ and then transform $\mZ$ to the cluster-distance space, i.e., we get a transformed matrix of size $N \times K$ where the entry in $i$-th row and $k$-th column is the euclidean distance of $\vz_i$ from $k$-th cluster center.
          A softmax over the columns of this matrix gives us initial probabilistic cluster assignments.
          We then pre-train the cluster embeddings $\vg_k$ by minimizing the KL-divergence between $p_{\theta}(\vc|\mZ)$ and the initialized cluster assignment probabilities.
          This KL-divergence term is used as a substitute for $L_\mathrm{c}$.
          All the other loss terms function the same as detailed in \secref{sec:variationalModule} and \secref{sec:contrastiveRegularizationModules}
    \item \textbf{Joint Training: }
          The end-to-end training of \ours is performed in this step to minimize the loss in \eqref{eq:abstractLoss}.
\end{enumerate}
The architecture is implemented in pytorch-geometric\cite{torchGeoetric} and the meta-path samples are generated using dgl\cite{dgl}.
For every meta-path, we select $10$ samples from every starting node.
For the loss $L^p$ in \eqref{eq:LpWithJNegativeSamples}, we select $J = 1$, i.e., we generate one negative sample for every positive sample.
The joint clustering assignments are obtained as the \texttt{argmax} of the corresponding cluster distribution.
All the results are presented as an average of $10$ runs.
Kindly refer to the supplementary material\cite{supplementary} for further details of implementation.

\subsection{Node Clustering}\label{sec:experimentsNodeClustering}
\begin{table}[]
    \resizebox{1\linewidth}{!}{%
                \begin{tabular}{l|cccc}
                    \toprule
                    \textbf{Method}                           & \textbf{DBLP-P} & \textbf{DBLP-A} & \textbf{ACM}   & \textbf{IMDB} \\
                    \midrule
                    DeepWalk                                  & 46.75           & 66.25           & 48.81          & 0.41          \\
                    LINE-1                                    & 42.18           & 29.98           & 37.75          & 0.03          \\
                    LINE-2                                    & 46.83           & 61.11           & 41.8           & 0.03          \\
                    GAE                                       & 63.21           & 65.43           & 41.03          & 2.91          \\
                    VGAE                                      & 62.76           & 63.42           & 42.14          & 3.51          \\
                    DGI                                       & 37.33           & 10.98           & 39.66          & 0.53          \\
                    GEMSEC                                    & 41.18           & 62.22           & 32.69          & 0.21          \\
                    CNRL                                      & 39.02           & 66.18           & 36.81          & 0.30          \\
                    CommunityGAN                              & 41.43           & 66.93           & 38.06          & 0.68          \\
                    \midrule
                    DHNE                                      & 35.33           & 21.00           & 20.25          & 0.05          \\
                    Metapath2Vec                              & 56.89           & 68.74           & 42.71          & 0.09          \\
                    HIN2Vec                                   & 30.47           & 65.79           & 42.28          & 0.04          \\
                    HERec                                     & 39.46           & 24.09           & 40.70          & 0.51          \\
                    HDGI                                      & 41.48           & 29.46           & 41.05          & 0.71          \\
                    NSHE                                      & 65.54           & 69.52           & 42.19          & 5.61          \\
                    HeGAN                                     & 60.78           & 68.95           & 43.35          & 6.56          \\
                    \midrule
                    \ours with $\vg_k$ and $L_{\mathrm{cont}}$     & \textbf{72.85}  & \textbf{71.25}  & \textbf{52.44} & 6.63          \\
                    \ours with $\vg_k$ without $L_{\mathrm{cont}}$ & 72.35           & 69.85           & 50.65          & \textbf{7.59} \\
                    \ours with KM and $L_{\mathrm{cont}}$        & 70.33           & 68.30           & 50.93          & 4.58          \\
                    \ours with KM without $L_{\mathrm{cont}}$      & 70.89           & 69.47           & 51.77          & 3.38          \\
                    \bottomrule
                \end{tabular}%
    }
    \caption{
        NMI scores on node clustering task.
        Best results for each dataset are bold.
        For comparison, we give the results both with and without contrastive loss $L_{\mathrm{cont}}$.
        For \ours results, We use $\vg_k$ when the results are obtained by utilizing cluster embeddings for inferring the cluster assignments.
        Moreover, we also give the results obtained by fitting K-Means (abbreviated as KM) on the learned embeddings.
    }
    \label{tab:resClustering}
\end{table}
\begin{table*}[]
    \resizebox{0.95\linewidth}{!}{%
        {\setlength{\tabcolsep}{1.2em}%
                \begin{tabular}{l|cccc|cccc}
                    \toprule
                    \multirow{3}{*}{\textbf{Method}} & \multicolumn{4}{c|}{\textbf{F1-Micro}} & \multicolumn{4}{c}{\textbf{F1-Micro}}                                                                                                         \\
                                                     & \cline{1-8}
                                                     & \textbf{DBLP-P}                        & \textbf{DBLP-A}                       & \textbf{ACM}   & \textbf{IMDB}  & \textbf{DBLP-P} & \textbf{DBLP-A} & \textbf{ACM}   & \textbf{IMDB}  \\
                    \midrule
                    DeepWalk                         & 90.12                                  & 89.44                                 & 82.17          & 56.52          & 89.45           & 88.48           & 81.82          & 55.24          \\
                    LINE-1                           & 81.43                                  & 82.32                                 & 82.46          & 43.75          & 80.74           & 80.20           & 82.35          & 39.87          \\
                    LINE-2                           & 84.76                                  & 88.76                                 & 82.21          & 40.54          & 83.45           & 87.35           & 81.32          & 33.06          \\
                    GAE                              & 90.09                                  & 77.54                                 & 81.97          & 55.16          & 89.23           & 60.21           & 81.47          & 53.57          \\
                    VGAE                             & 90.95                                  & 71.43                                 & 81.59          & 57.47          & 90.04           & 69.03           & 81.30          & 56.06          \\
                    DGI                              & 95.24                                  & 91.81                                 & 81.65          & 58.42          & 94.51           & 91.20           & 81.79          & 56.94          \\
                    GEMSEC                           & 78.38                                  & 87.45                                 & 62.05          & 51.56          & 77.85           & 87.01           & 61.24          & 50.20          \\
                    CNRL                             & 81.43                                  & 83.20                                 & 66.58          & 43.12          & 80.83           & 82.68           & 65.28          & 41.48          \\
                    CommunityGAN                     & 72.01                                  & 67.81                                 & 52.87          & 33.81          & 66.54           & 71.55           & 51.67          & 31.09          \\
                    \midrule
                    DHNE                             & 85.71                                  & 73.30                                 & 65.27          & 38.99          & 84.67           & 67.61           & 62.31          & 30.53          \\
                    Metapath2Vec                     & 92.86                                  & 89.36                                 & 83.60          & 51.90          & 92.44           & 87.95           & 82.77          & 50.21          \\
                    HIN2Vec                          & 83.81                                  & 90.30                                 & 54.30          & 48.02          & 83.85           & 89.46           & 48.59          & 46.24          \\
                    HERec                            & 90.47                                  & 86.21                                 & 81.89          & 54.48          & 87.50           & 84.55           & 81.74          & 53.46          \\
                    HDGI                             & 95.24                                  & 92.27                                 & 82.06          & 57.81          & 94.51           & 91.81           & 81.64          & 55.73          \\
                    NSHE                             & 95.24                                  & 93.10                                 & 82.52          & 59.21          & 94.76           & 92.37           & 82.67          & 58.31          \\
                    HeGAN                            & 88.79                                  & 90.48                                 & 83.09          & 58.56          & 83.81           & 89.27           & 82.94          & 57.12          \\
                    \midrule
                    \ours                            & \textbf{100.00}                        & \textbf{93.57}                        & \textbf{83.70} & \textbf{60.73} & \textbf{100.00} & \textbf{92.80}  & \textbf{83.48} & \textbf{59.28} \\
                    \ours without $L_{\mathrm{cont}}$     & \textbf{100.00}                        & 93.10                                 & 80.46          & 57.60          & \textbf{100.00} & 92.22           & 79.85          & 53.27          \\
                    \bottomrule
                \end{tabular}%
            }
    }
    \captionsetup{width=0.95\linewidth}
    \caption{
        F1-Micro and F1-Macro scores on node classification task.
        Best results for each dataset are bold.
        For comparison, we give the results both with and without contrastive loss $L_{\mathrm{cont}}$.
        It can be readily observed that \ours outperforms the baselines for all the datasets.
        Moreover, the results are always improved by including the contrastive modules.
    }
    \label{tab:resClassification}
\end{table*}
We start by evaluating the learned embeddings on clustering.
Apart from GEMSEC, CNRL, and CommunityGAN, no baseline learns clustering assignments jointly with the embeddings.
Therefore we choose K-Means to find the cluster assignments for such algorithms.
For \ours we give the results both with and without $L_{\mathrm{cont}}$.
Moreover, for comparison with the baselines, we also look at the case where the cluster embeddings $\vg_k$ are used during joint training but not used for extracting the cluster assignments.
Instead, we use the assignments obtained by directly fitting K-Means on the learned HIN embedding.
We use normalized mutual information (NMI) score for the quantitative evaluation of the performance.

\Tabref{tab:resClustering} gives the comparison of different algorithms for node clustering.
The results in the table are divided into three sections i.e., the approaches dealing with homogeneous networks, the approaches for unsupervised HIN embedding, and different variants of the proposed method.
\ours results are the best on all the datasets.
In addition, the use of $L_{\mathrm{cont}}$ improves the results in $3$ out of $4$ cases, the exception being the IMDB dataset.
It is rather difficult to comment on the reason behind this, because this dataset is inherently not easy to cluster as demonstrated by the low NMI scores for all the algorithms.
Whether or not we use $L_{\mathrm{cont}}$, the performance achieved by using the jointly learned cluster assignments is always better than the one where K-Means is used to cluster the HIN embedding.
So joint learning of cluster assignments and network embedding consistently outperforms the case where these two tasks are treated independently.
This conforms to the results obtained in the domain of euclidean data and homogeneous networks.
It also highlights the efficacy of the variational module and validates the intuition behind the choices of involved distributions.
Moreover, as shown by the last two rows of \tabref{tab:resClustering}, $L_{\mathrm{cont}}$ has a negative effect on the performance in $3$ out of $4$ datasets if K-Means is used to get cluster assignments.
Apart from \ours, the HIN embedding methods, e.g., HeGAN, NSHE, and Metapath2Vec, generally perform better than the approaches that treat a HIN as a homogeneous graph.

\subsection{Node Classification}
For node classification, we first learn the HIN embedding in an unsupervised manner for each of the four cases.
Afterward, $80\%$ of the labels are used for training the logistic classifier using \texttt{lbfgs} solver\cite{lbfgs,scikit-learn}.
We keep the split the same as our competitors for a fair comparison.
The performance is evaluated using F1-Micro and F1-macro scores.
\Tabref{tab:resClassification} gives a comparison of \ours with the baselines.
In all the datasets, \ours gives the best performance.
However, the performance suffers when $L_{\mathrm{cont}}$ is ignored, thereby providing empirical evidence to the utility of the contrastive modules for improving HIN embedding quality.
The improvement margin is maximum in case of DBLP-P, because relatively fewer labeled nodes are available for this dataset.
So, compared to the competitors, correctly classifying even a few more nodes enables us to achieve perfect results.
Among the methods that jointly learn homogeneous network embedding and clustering assignments, CommunityGAN performs poorly in particular.
A possible reason is restricting the latent dimensions to be the same as the number of clusters.
While it can yield acceptable clusters, it makes the downstream node classification difficult for \textit{linear} classifiers, especially when $K$ is small.
An interesting observation lies in the results of contrastive approaches (DGI and HDGI) and autoencoder based models in \tabref{tab:resClustering} and \tabref{tab:resClassification}.
For clustering in \tabref{tab:resClustering}, DGI/HDGI results are usually quite poor compared to GAE and VGAE.
However, for classification in \tabref{tab:resClassification}, DGI and HDGI almost always outperform GAE and VGAE.
This gives a hint that an approach based on edge reconstruction might be better suited for HIN clustering, whereas a contrastive approach could help in the general improvement of node embedding quality as evaluated by the downstream node classification.
Since \ours makes use of a variational module (aimed to reconstruct $\mA$) as well as the contrastive modules, it performs well in both tasks.

\section{Conclusion}
We make the first attempt at joint learning of cluster assignments and HIN embedding.
This is achieved by refining a single target HIN embedding using a variational module and multiple contrastive modules.
The variational module aims at reconstruction of the adjacency matrix $\mA$ in a cluster-aware manner.
The contrastive modules attempt to preserve the high-order HIN structure.
Specifically, every contrastive module is assigned the task of distinguishing between positive and negative/corrupted samples of a certain meta-path.
The joint training of the variational and contrastive modules yields the HIN embedding that leverages the local information (provided by the pairwise relations) as well as the global information (present in cluster assignments and high-order semantics as dictated by the samples of different meta-paths).
In addition to the HIN embedding, \ours simultaneously learns the cluster embeddings and consequently the cluster assignments for HIN nodes.
These jointly learned cluster assignments consistently outperform the basic clustering strategy commonly used for HINs, i.e., applying some off-the-shelf clustering algorithm (e.g., K-Means or GMM, etc.) to the learned HIN embedding.
Moreover, the HIN embedding is also efficient for downstream node classification task, as demonstrated by comparison with many competitive baselines.

\begin{acks}
  This work has been supported by the Bavarian Ministry of Economic Affairs, Regional Development and Energy through the WoWNet project IUK-1902-003// IUK625/002.
\end{acks}

\bibliographystyle{ACM-Reference-Format}
\bibliography{main}


\begin{thebibliography}{63}


\ifx \showCODEN    \undefined \def \showCODEN     #1{\unskip}     \fi
\ifx \showDOI      \undefined \def \showDOI       #1{#1}\fi
\ifx \showISBNx    \undefined \def \showISBNx     #1{\unskip}     \fi
\ifx \showISBNxiii \undefined \def \showISBNxiii  #1{\unskip}     \fi
\ifx \showISSN     \undefined \def \showISSN      #1{\unskip}     \fi
\ifx \showLCCN     \undefined \def \showLCCN      #1{\unskip}     \fi
\ifx \shownote     \undefined \def \shownote      #1{#1}          \fi
\ifx \showarticletitle \undefined \def \showarticletitle #1{#1}   \fi
\ifx \showURL      \undefined \def \showURL       {\relax}        \fi
\providecommand\bibfield[2]{#2}
\providecommand\bibinfo[2]{#2}
\providecommand\natexlab[1]{#1}
\providecommand\showeprint[2][]{arXiv:#2}

\bibitem[\protect\citeauthoryear{Anonymous}{Anonymous}{2021}]%
        {supplementary}
\bibfield{author}{\bibinfo{person}{Anonymous}.}
  \bibinfo{year}{2021}\natexlab{}.
\newblock \bibinfo{title}{Supplementary Material pdf And Implementation Code}.
\newblock
  \bibinfo{howpublished}{\url{https://drive.google.com/drive/folders/1eRuurgt3knR3bqhXPy7pT5QRXdX_GAMQ?usp=sharing}}.
\newblock


\bibitem[\protect\citeauthoryear{Ata, Fang, Wu, Li, and Xiao}{Ata
  et~al\mbox{.}}{2017}]%
        {applications-bio1}
\bibfield{author}{\bibinfo{person}{Sezin~Kircali Ata}, \bibinfo{person}{Yuan
  Fang}, \bibinfo{person}{Min Wu}, \bibinfo{person}{Xiao-Li Li}, {and}
  \bibinfo{person}{Xiaokui Xiao}.} \bibinfo{year}{2017}\natexlab{}.
\newblock \showarticletitle{Disease gene classification with metagraph
  representations}.
\newblock \bibinfo{journal}{\emph{Methods}}  \bibinfo{volume}{131}
  (\bibinfo{year}{2017}), \bibinfo{pages}{83--92}.
\newblock


\bibitem[\protect\citeauthoryear{Cai, Zheng, and Chang}{Cai
  et~al\mbox{.}}{2018a}]%
        {applications-survey2}
\bibfield{author}{\bibinfo{person}{Hongyun Cai}, \bibinfo{person}{Vincent~W
  Zheng}, {and} \bibinfo{person}{Kevin Chen-Chuan Chang}.}
  \bibinfo{year}{2018}\natexlab{a}.
\newblock \showarticletitle{A comprehensive survey of graph embedding:
  Problems, techniques, and applications}.
\newblock \bibinfo{journal}{\emph{IEEE Transactions on Knowledge and Data
  Engineering}} \bibinfo{volume}{30}, \bibinfo{number}{9}
  (\bibinfo{year}{2018}), \bibinfo{pages}{1616--1637}.
\newblock


\bibitem[\protect\citeauthoryear{Cai, Zheng, and Chang}{Cai
  et~al\mbox{.}}{2018b}]%
        {embeddings-survey1}
\bibfield{author}{\bibinfo{person}{Hongyun Cai}, \bibinfo{person}{Vincent~W
  Zheng}, {and} \bibinfo{person}{Kevin Chen-Chuan Chang}.}
  \bibinfo{year}{2018}\natexlab{b}.
\newblock \showarticletitle{A comprehensive survey of graph embedding:
  Problems, techniques, and applications}.
\newblock \bibinfo{journal}{\emph{IEEE Transactions on Knowledge and Data
  Engineering}} \bibinfo{volume}{30}, \bibinfo{number}{9}
  (\bibinfo{year}{2018}), \bibinfo{pages}{1616--1637}.
\newblock


\bibitem[\protect\citeauthoryear{Chami, Wolf, Juan, Sala, Ravi, and
  R{\'e}}{Chami et~al\mbox{.}}{2020}]%
        {applications-knowledgeGraphs2}
\bibfield{author}{\bibinfo{person}{Ines Chami}, \bibinfo{person}{Adva Wolf},
  \bibinfo{person}{Da-Cheng Juan}, \bibinfo{person}{Frederic Sala},
  \bibinfo{person}{Sujith Ravi}, {and} \bibinfo{person}{Christopher R{\'e}}.}
  \bibinfo{year}{2020}\natexlab{}.
\newblock \showarticletitle{Low-dimensional hyperbolic knowledge graph
  embeddings}.
\newblock \bibinfo{journal}{\emph{arXiv preprint arXiv:2005.00545}}
  (\bibinfo{year}{2020}).
\newblock


\bibitem[\protect\citeauthoryear{Chen and Sun}{Chen and Sun}{2017}]%
        {related-meta-path-fusion2}
\bibfield{author}{\bibinfo{person}{Ting Chen} {and} \bibinfo{person}{Yizhou
  Sun}.} \bibinfo{year}{2017}\natexlab{}.
\newblock \showarticletitle{Task-guided and path-augmented heterogeneous
  network embedding for author identification}. In
  \bibinfo{booktitle}{\emph{Proceedings of the Tenth ACM International
  Conference on Web Search and Data Mining}}. \bibinfo{pages}{295--304}.
\newblock


\bibitem[\protect\citeauthoryear{Cui, Wang, Pei, and Zhu}{Cui
  et~al\mbox{.}}{2018}]%
        {embeddings-survey2}
\bibfield{author}{\bibinfo{person}{Peng Cui}, \bibinfo{person}{Xiao Wang},
  \bibinfo{person}{Jian Pei}, {and} \bibinfo{person}{Wenwu Zhu}.}
  \bibinfo{year}{2018}\natexlab{}.
\newblock \showarticletitle{A survey on network embedding}.
\newblock \bibinfo{journal}{\emph{IEEE Transactions on Knowledge and Data
  Engineering}} \bibinfo{volume}{31}, \bibinfo{number}{5}
  (\bibinfo{year}{2018}), \bibinfo{pages}{833--852}.
\newblock


\bibitem[\protect\citeauthoryear{Dilokthanakul, Mediano, Garnelo, Lee,
  Salimbeni, Arulkumaran, and Shanahan}{Dilokthanakul et~al\mbox{.}}{2016}]%
        {euclidean-joint5}
\bibfield{author}{\bibinfo{person}{Nat Dilokthanakul},
  \bibinfo{person}{Pedro~AM Mediano}, \bibinfo{person}{Marta Garnelo},
  \bibinfo{person}{Matthew~CH Lee}, \bibinfo{person}{Hugh Salimbeni},
  \bibinfo{person}{Kai Arulkumaran}, {and} \bibinfo{person}{Murray Shanahan}.}
  \bibinfo{year}{2016}\natexlab{}.
\newblock \showarticletitle{Deep unsupervised clustering with gaussian mixture
  variational autoencoders}.
\newblock \bibinfo{journal}{\emph{arXiv preprint arXiv:1611.02648}}
  (\bibinfo{year}{2016}).
\newblock


\bibitem[\protect\citeauthoryear{Doersch}{Doersch}{2016}]%
        {vae-tutorial}
\bibfield{author}{\bibinfo{person}{Carl Doersch}.}
  \bibinfo{year}{2016}\natexlab{}.
\newblock \showarticletitle{Tutorial on variational autoencoders}.
\newblock \bibinfo{journal}{\emph{arXiv preprint arXiv:1606.05908}}
  (\bibinfo{year}{2016}).
\newblock


\bibitem[\protect\citeauthoryear{Dong, Chawla, and Swami}{Dong
  et~al\mbox{.}}{2017}]%
        {related:metapath2vec}
\bibfield{author}{\bibinfo{person}{Yuxiao Dong}, \bibinfo{person}{Nitesh~V
  Chawla}, {and} \bibinfo{person}{Ananthram Swami}.}
  \bibinfo{year}{2017}\natexlab{}.
\newblock \showarticletitle{metapath2vec: Scalable representation learning for
  heterogeneous networks}. In \bibinfo{booktitle}{\emph{Proceedings of the 23rd
  ACM SIGKDD international conference on knowledge discovery and data mining}}.
  \bibinfo{pages}{135--144}.
\newblock


\bibitem[\protect\citeauthoryear{Duvenaud, Maclaurin, Aguilera-Iparraguirre,
  G{\'o}mez-Bombarelli, Hirzel, Aspuru-Guzik, and Adams}{Duvenaud
  et~al\mbox{.}}{2015}]%
        {applications-chem2}
\bibfield{author}{\bibinfo{person}{David Duvenaud}, \bibinfo{person}{Dougal
  Maclaurin}, \bibinfo{person}{Jorge Aguilera-Iparraguirre},
  \bibinfo{person}{Rafael G{\'o}mez-Bombarelli}, \bibinfo{person}{Timothy
  Hirzel}, \bibinfo{person}{Al{\'a}n Aspuru-Guzik}, {and}
  \bibinfo{person}{Ryan~P Adams}.} \bibinfo{year}{2015}\natexlab{}.
\newblock \showarticletitle{Convolutional networks on graphs for learning
  molecular fingerprints}.
\newblock \bibinfo{journal}{\emph{arXiv preprint arXiv:1509.09292}}
  (\bibinfo{year}{2015}).
\newblock


\bibitem[\protect\citeauthoryear{Fey and Lenssen}{Fey and Lenssen}{2019}]%
        {torchGeoetric}
\bibfield{author}{\bibinfo{person}{Matthias Fey} {and} \bibinfo{person}{Jan~E.
  Lenssen}.} \bibinfo{year}{2019}\natexlab{}.
\newblock \showarticletitle{Fast Graph Representation Learning with {PyTorch
  Geometric}}. In \bibinfo{booktitle}{\emph{ICLR Workshop on Representation
  Learning on Graphs and Manifolds}}.
\newblock


\bibitem[\protect\citeauthoryear{Fout}{Fout}{2017}]%
        {applications-bio2}
\bibfield{author}{\bibinfo{person}{Alex~M Fout}.}
  \bibinfo{year}{2017}\natexlab{}.
\newblock \emph{\bibinfo{title}{Protein interface prediction using graph
  convolutional networks}}.
\newblock \bibinfo{thesistype}{Ph.D. Dissertation}. \bibinfo{school}{Colorado
  State University}.
\newblock


\bibitem[\protect\citeauthoryear{Fu, Lee, and Lei}{Fu et~al\mbox{.}}{2017}]%
        {related:Hin2vec}
\bibfield{author}{\bibinfo{person}{Tao-yang Fu}, \bibinfo{person}{Wang-Chien
  Lee}, {and} \bibinfo{person}{Zhen Lei}.} \bibinfo{year}{2017}\natexlab{}.
\newblock \showarticletitle{Hin2vec: Explore meta-paths in heterogeneous
  information networks for representation learning}. In
  \bibinfo{booktitle}{\emph{Proceedings of the 2017 ACM on Conference on
  Information and Knowledge Management}}. \bibinfo{pages}{1797--1806}.
\newblock


\bibitem[\protect\citeauthoryear{Gilmer, Schoenholz, Riley, Vinyals, and
  Dahl}{Gilmer et~al\mbox{.}}{2017}]%
        {applications-chem1}
\bibfield{author}{\bibinfo{person}{Justin Gilmer}, \bibinfo{person}{Samuel~S
  Schoenholz}, \bibinfo{person}{Patrick~F Riley}, \bibinfo{person}{Oriol
  Vinyals}, {and} \bibinfo{person}{George~E Dahl}.}
  \bibinfo{year}{2017}\natexlab{}.
\newblock \showarticletitle{Neural message passing for quantum chemistry}. In
  \bibinfo{booktitle}{\emph{International Conference on Machine Learning}}.
  PMLR, \bibinfo{pages}{1263--1272}.
\newblock


\bibitem[\protect\citeauthoryear{Griffa, Ricaud, Benzi, Bresson, Daducci,
  Vandergheynst, Thiran, and Hagmann}{Griffa et~al\mbox{.}}{2017}]%
        {applications-bio3}
\bibfield{author}{\bibinfo{person}{Alessandra Griffa},
  \bibinfo{person}{Benjamin Ricaud}, \bibinfo{person}{Kirell Benzi},
  \bibinfo{person}{Xavier Bresson}, \bibinfo{person}{Alessandro Daducci},
  \bibinfo{person}{Pierre Vandergheynst}, \bibinfo{person}{Jean-Philippe
  Thiran}, {and} \bibinfo{person}{Patric Hagmann}.}
  \bibinfo{year}{2017}\natexlab{}.
\newblock \showarticletitle{Transient networks of spatio-temporal connectivity
  map communication pathways in brain functional systems}.
\newblock \bibinfo{journal}{\emph{NeuroImage}}  \bibinfo{volume}{155}
  (\bibinfo{year}{2017}), \bibinfo{pages}{490--502}.
\newblock


\bibitem[\protect\citeauthoryear{Grover and Leskovec}{Grover and
  Leskovec}{2016}]%
        {related:node2vec}
\bibfield{author}{\bibinfo{person}{Aditya Grover} {and} \bibinfo{person}{Jure
  Leskovec}.} \bibinfo{year}{2016}\natexlab{}.
\newblock \showarticletitle{node2vec: Scalable feature learning for networks}.
  In \bibinfo{booktitle}{\emph{Proceedings of the 22nd ACM SIGKDD international
  conference on Knowledge discovery and data mining}}.
  \bibinfo{pages}{855--864}.
\newblock


\bibitem[\protect\citeauthoryear{Hamilton, Ying, and Leskovec}{Hamilton
  et~al\mbox{.}}{2017}]%
        {graph-sage}
\bibfield{author}{\bibinfo{person}{Will Hamilton}, \bibinfo{person}{Zhitao
  Ying}, {and} \bibinfo{person}{Jure Leskovec}.}
  \bibinfo{year}{2017}\natexlab{}.
\newblock \showarticletitle{Inductive representation learning on large graphs}.
  In \bibinfo{booktitle}{\emph{Advances in neural information processing
  systems}}. \bibinfo{pages}{1024--1034}.
\newblock


\bibitem[\protect\citeauthoryear{Hjelm, Fedorov, Lavoie-Marchildon, Grewal,
  Bachman, Trischler, and Bengio}{Hjelm et~al\mbox{.}}{2018a}]%
        {deep-infomax}
\bibfield{author}{\bibinfo{person}{R~Devon Hjelm}, \bibinfo{person}{Alex
  Fedorov}, \bibinfo{person}{Samuel Lavoie-Marchildon}, \bibinfo{person}{Karan
  Grewal}, \bibinfo{person}{Phil Bachman}, \bibinfo{person}{Adam Trischler},
  {and} \bibinfo{person}{Yoshua Bengio}.} \bibinfo{year}{2018}\natexlab{a}.
\newblock \showarticletitle{Learning deep representations by mutual information
  estimation and maximization}.
\newblock \bibinfo{journal}{\emph{arXiv preprint arXiv:1808.06670}}
  (\bibinfo{year}{2018}).
\newblock


\bibitem[\protect\citeauthoryear{Hjelm, Fedorov, Lavoie-Marchildon, Grewal,
  Bachman, Trischler, and Bengio}{Hjelm et~al\mbox{.}}{2018b}]%
        {deepinfomax}
\bibfield{author}{\bibinfo{person}{R~Devon Hjelm}, \bibinfo{person}{Alex
  Fedorov}, \bibinfo{person}{Samuel Lavoie-Marchildon}, \bibinfo{person}{Karan
  Grewal}, \bibinfo{person}{Phil Bachman}, \bibinfo{person}{Adam Trischler},
  {and} \bibinfo{person}{Yoshua Bengio}.} \bibinfo{year}{2018}\natexlab{b}.
\newblock \showarticletitle{Learning deep representations by mutual information
  estimation and maximization}.
\newblock \bibinfo{journal}{\emph{arXiv preprint arXiv:1808.06670}}
  (\bibinfo{year}{2018}).
\newblock


\bibitem[\protect\citeauthoryear{Hu, Fang, and Shi}{Hu et~al\mbox{.}}{2019}]%
        {related:hegan}
\bibfield{author}{\bibinfo{person}{Binbin Hu}, \bibinfo{person}{Yuan Fang},
  {and} \bibinfo{person}{Chuan Shi}.} \bibinfo{year}{2019}\natexlab{}.
\newblock \showarticletitle{Adversarial learning on heterogeneous information
  networks}. In \bibinfo{booktitle}{\emph{Proceedings of the 25th ACM SIGKDD
  International Conference on Knowledge Discovery \& Data Mining}}.
  \bibinfo{pages}{120--129}.
\newblock


\bibitem[\protect\citeauthoryear{Huang, Huang, Wang, and Wang}{Huang
  et~al\mbox{.}}{2014}]%
        {euclidean-joint3}
\bibfield{author}{\bibinfo{person}{Peihao Huang}, \bibinfo{person}{Yan Huang},
  \bibinfo{person}{Wei Wang}, {and} \bibinfo{person}{Liang Wang}.}
  \bibinfo{year}{2014}\natexlab{}.
\newblock \showarticletitle{Deep embedding network for clustering}. In
  \bibinfo{booktitle}{\emph{2014 22nd International conference on pattern
  recognition}}. IEEE, \bibinfo{pages}{1532--1537}.
\newblock


\bibitem[\protect\citeauthoryear{Huang and Mamoulis}{Huang and
  Mamoulis}{2017}]%
        {related-meta-path-fusion1}
\bibfield{author}{\bibinfo{person}{Zhipeng Huang} {and} \bibinfo{person}{Nikos
  Mamoulis}.} \bibinfo{year}{2017}\natexlab{}.
\newblock \showarticletitle{Heterogeneous information network embedding for
  meta path based proximity}.
\newblock \bibinfo{journal}{\emph{arXiv preprint arXiv:1701.05291}}
  (\bibinfo{year}{2017}).
\newblock


\bibitem[\protect\citeauthoryear{Hussein, Yang, and Cudr{\'e}-Mauroux}{Hussein
  et~al\mbox{.}}{2018}]%
        {related-just}
\bibfield{author}{\bibinfo{person}{Rana Hussein}, \bibinfo{person}{Dingqi
  Yang}, {and} \bibinfo{person}{Philippe Cudr{\'e}-Mauroux}.}
  \bibinfo{year}{2018}\natexlab{}.
\newblock \showarticletitle{Are meta-paths necessary? Revisiting heterogeneous
  graph embeddings}. In \bibinfo{booktitle}{\emph{Proceedings of the 27th ACM
  International Conference on Information and Knowledge Management}}.
  \bibinfo{pages}{437--446}.
\newblock


\bibitem[\protect\citeauthoryear{Jang, Gu, and Poole}{Jang
  et~al\mbox{.}}{2016}]%
        {gumbel-softmax}
\bibfield{author}{\bibinfo{person}{Eric Jang}, \bibinfo{person}{Shixiang Gu},
  {and} \bibinfo{person}{Ben Poole}.} \bibinfo{year}{2016}\natexlab{}.
\newblock \showarticletitle{Categorical reparameterization with
  gumbel-softmax}.
\newblock \bibinfo{journal}{\emph{arXiv preprint arXiv:1611.01144}}
  (\bibinfo{year}{2016}).
\newblock


\bibitem[\protect\citeauthoryear{Jia, Zhang, Zhang, and Wang}{Jia
  et~al\mbox{.}}{2019}]%
        {jia2019communitygan}
\bibfield{author}{\bibinfo{person}{Yuting Jia}, \bibinfo{person}{Qinqin Zhang},
  \bibinfo{person}{Weinan Zhang}, {and} \bibinfo{person}{Xinbing Wang}.}
  \bibinfo{year}{2019}\natexlab{}.
\newblock \showarticletitle{CommunityGAN: Community detection with generative
  adversarial nets}. In \bibinfo{booktitle}{\emph{The World Wide Web
  Conference}}. \bibinfo{pages}{784--794}.
\newblock


\bibitem[\protect\citeauthoryear{Jiang, Zheng, Tan, Tang, and Zhou}{Jiang
  et~al\mbox{.}}{2016}]%
        {euclidean-joint4}
\bibfield{author}{\bibinfo{person}{Zhuxi Jiang}, \bibinfo{person}{Yin Zheng},
  \bibinfo{person}{Huachun Tan}, \bibinfo{person}{Bangsheng Tang}, {and}
  \bibinfo{person}{Hanning Zhou}.} \bibinfo{year}{2016}\natexlab{}.
\newblock \showarticletitle{Variational deep embedding: An unsupervised and
  generative approach to clustering}.
\newblock \bibinfo{journal}{\emph{arXiv preprint arXiv:1611.05148}}
  (\bibinfo{year}{2016}).
\newblock


\bibitem[\protect\citeauthoryear{Khan, Anwaar, and Kleinsteuber}{Khan
  et~al\mbox{.}}{2020}]%
        {khan2020epitomic}
\bibfield{author}{\bibinfo{person}{Rayyan~Ahmad Khan},
  \bibinfo{person}{Muhammad~Umer Anwaar}, {and} \bibinfo{person}{Martin
  Kleinsteuber}.} \bibinfo{year}{2020}\natexlab{}.
\newblock \bibinfo{title}{Epitomic Variational Graph Autoencoder}.
\newblock
\newblock
\showeprint[arxiv]{2004.01468}~[cs.LG]


\bibitem[\protect\citeauthoryear{Kingma and Welling}{Kingma and
  Welling}{2013}]%
        {vae-kingma}
\bibfield{author}{\bibinfo{person}{Diederik~P Kingma} {and}
  \bibinfo{person}{Max Welling}.} \bibinfo{year}{2013}\natexlab{}.
\newblock \showarticletitle{Auto-encoding variational bayes}.
\newblock \bibinfo{journal}{\emph{arXiv preprint arXiv:1312.6114}}
  (\bibinfo{year}{2013}).
\newblock


\bibitem[\protect\citeauthoryear{Kipf and Welling}{Kipf and Welling}{2016a}]%
        {exp:split-2}
\bibfield{author}{\bibinfo{person}{Thomas~N Kipf} {and} \bibinfo{person}{Max
  Welling}.} \bibinfo{year}{2016}\natexlab{a}.
\newblock \showarticletitle{Semi-supervised classification with graph
  convolutional networks}.
\newblock \bibinfo{journal}{\emph{arXiv preprint arXiv:1609.02907}}
  (\bibinfo{year}{2016}).
\newblock


\bibitem[\protect\citeauthoryear{Kipf and Welling}{Kipf and Welling}{2016b}]%
        {vgae}
\bibfield{author}{\bibinfo{person}{Thomas~N Kipf} {and} \bibinfo{person}{Max
  Welling}.} \bibinfo{year}{2016}\natexlab{b}.
\newblock \showarticletitle{Variational graph auto-encoders}.
\newblock \bibinfo{journal}{\emph{arXiv preprint arXiv:1611.07308}}
  (\bibinfo{year}{2016}).
\newblock


\bibitem[\protect\citeauthoryear{Lu, Shi, Hu, and Liu}{Lu
  et~al\mbox{.}}{2019}]%
        {related-rhine}
\bibfield{author}{\bibinfo{person}{Yuanfu Lu}, \bibinfo{person}{Chuan Shi},
  \bibinfo{person}{Linmei Hu}, {and} \bibinfo{person}{Zhiyuan Liu}.}
  \bibinfo{year}{2019}\natexlab{}.
\newblock \showarticletitle{Relation structure-aware heterogeneous information
  network embedding}. In \bibinfo{booktitle}{\emph{Proceedings of the AAAI
  Conference on Artificial Intelligence}}, Vol.~\bibinfo{volume}{33}.
  \bibinfo{pages}{4456--4463}.
\newblock


\bibitem[\protect\citeauthoryear{Monti, Bronstein, and Bresson}{Monti
  et~al\mbox{.}}{2017}]%
        {applications-social2}
\bibfield{author}{\bibinfo{person}{Federico Monti}, \bibinfo{person}{Michael~M
  Bronstein}, {and} \bibinfo{person}{Xavier Bresson}.}
  \bibinfo{year}{2017}\natexlab{}.
\newblock \showarticletitle{Geometric matrix completion with recurrent
  multi-graph neural networks}.
\newblock \bibinfo{journal}{\emph{arXiv preprint arXiv:1704.06803}}
  (\bibinfo{year}{2017}).
\newblock


\bibitem[\protect\citeauthoryear{Monti, Frasca, Eynard, Mannion, and
  Bronstein}{Monti et~al\mbox{.}}{2019}]%
        {applications-social4}
\bibfield{author}{\bibinfo{person}{Federico Monti}, \bibinfo{person}{Fabrizio
  Frasca}, \bibinfo{person}{Davide Eynard}, \bibinfo{person}{Damon Mannion},
  {and} \bibinfo{person}{Michael~M Bronstein}.}
  \bibinfo{year}{2019}\natexlab{}.
\newblock \showarticletitle{Fake news detection on social media using geometric
  deep learning}.
\newblock \bibinfo{journal}{\emph{arXiv preprint arXiv:1902.06673}}
  (\bibinfo{year}{2019}).
\newblock


\bibitem[\protect\citeauthoryear{Pedregosa, Varoquaux, Gramfort, Michel,
  Thirion, Grisel, Blondel, Prettenhofer, Weiss, Dubourg, Vanderplas, Passos,
  Cournapeau, Brucher, Perrot, and Duchesnay}{Pedregosa et~al\mbox{.}}{2011}]%
        {scikit-learn}
\bibfield{author}{\bibinfo{person}{F. Pedregosa}, \bibinfo{person}{G.
  Varoquaux}, \bibinfo{person}{A. Gramfort}, \bibinfo{person}{V. Michel},
  \bibinfo{person}{B. Thirion}, \bibinfo{person}{O. Grisel},
  \bibinfo{person}{M. Blondel}, \bibinfo{person}{P. Prettenhofer},
  \bibinfo{person}{R. Weiss}, \bibinfo{person}{V. Dubourg}, \bibinfo{person}{J.
  Vanderplas}, \bibinfo{person}{A. Passos}, \bibinfo{person}{D. Cournapeau},
  \bibinfo{person}{M. Brucher}, \bibinfo{person}{M. Perrot}, {and}
  \bibinfo{person}{E. Duchesnay}.} \bibinfo{year}{2011}\natexlab{}.
\newblock \showarticletitle{Scikit-learn: Machine Learning in {P}ython}.
\newblock \bibinfo{journal}{\emph{Journal of Machine Learning Research}}
  \bibinfo{volume}{12} (\bibinfo{year}{2011}), \bibinfo{pages}{2825--2830}.
\newblock


\bibitem[\protect\citeauthoryear{Perozzi, Al-Rfou, and Skiena}{Perozzi
  et~al\mbox{.}}{2014}]%
        {related:deepwalk}
\bibfield{author}{\bibinfo{person}{Bryan Perozzi}, \bibinfo{person}{Rami
  Al-Rfou}, {and} \bibinfo{person}{Steven Skiena}.}
  \bibinfo{year}{2014}\natexlab{}.
\newblock \showarticletitle{Deepwalk: Online learning of social
  representations}. In \bibinfo{booktitle}{\emph{Proceedings of the 20th ACM
  SIGKDD international conference on Knowledge discovery and data mining}}.
  \bibinfo{pages}{701--710}.
\newblock


\bibitem[\protect\citeauthoryear{Ren, Liu, Huang, Dai, Bo, and Zhang}{Ren
  et~al\mbox{.}}{2019}]%
        {related-hetero-dgi}
\bibfield{author}{\bibinfo{person}{Yuxiang Ren}, \bibinfo{person}{Bo Liu},
  \bibinfo{person}{Chao Huang}, \bibinfo{person}{Peng Dai},
  \bibinfo{person}{Liefeng Bo}, {and} \bibinfo{person}{Jiawei Zhang}.}
  \bibinfo{year}{2019}\natexlab{}.
\newblock \showarticletitle{Heterogeneous deep graph infomax}.
\newblock \bibinfo{journal}{\emph{arXiv preprint arXiv:1911.08538}}
  (\bibinfo{year}{2019}).
\newblock


\bibitem[\protect\citeauthoryear{Ribeiro, Saverese, and Figueiredo}{Ribeiro
  et~al\mbox{.}}{2017}]%
        {ribeiro2017struc2vec}
\bibfield{author}{\bibinfo{person}{Leonardo~FR Ribeiro},
  \bibinfo{person}{Pedro~HP Saverese}, {and} \bibinfo{person}{Daniel~R
  Figueiredo}.} \bibinfo{year}{2017}\natexlab{}.
\newblock \showarticletitle{struc2vec: Learning node representations from
  structural identity}. In \bibinfo{booktitle}{\emph{Proceedings of the 23rd
  ACM SIGKDD international conference on knowledge discovery and data mining}}.
  \bibinfo{pages}{385--394}.
\newblock


\bibitem[\protect\citeauthoryear{Rozemberczki, Davies, Sarkar, and
  Sutton}{Rozemberczki et~al\mbox{.}}{2019}]%
        {gemsec}
\bibfield{author}{\bibinfo{person}{Benedek Rozemberczki}, \bibinfo{person}{Ryan
  Davies}, \bibinfo{person}{Rik Sarkar}, {and} \bibinfo{person}{Charles
  Sutton}.} \bibinfo{year}{2019}\natexlab{}.
\newblock \showarticletitle{Gemsec: Graph embedding with self clustering}. In
  \bibinfo{booktitle}{\emph{Proceedings of the 2019 IEEE/ACM international
  conference on advances in social networks analysis and mining}}.
  \bibinfo{pages}{65--72}.
\newblock


\bibitem[\protect\citeauthoryear{Schlichtkrull, Kipf, Bloem, Van Den~Berg,
  Titov, and Welling}{Schlichtkrull et~al\mbox{.}}{2018a}]%
        {applications-knowledgeGraphs1}
\bibfield{author}{\bibinfo{person}{Michael Schlichtkrull},
  \bibinfo{person}{Thomas~N Kipf}, \bibinfo{person}{Peter Bloem},
  \bibinfo{person}{Rianne Van Den~Berg}, \bibinfo{person}{Ivan Titov}, {and}
  \bibinfo{person}{Max Welling}.} \bibinfo{year}{2018}\natexlab{a}.
\newblock \showarticletitle{Modeling relational data with graph convolutional
  networks}. In \bibinfo{booktitle}{\emph{European semantic web conference}}.
  Springer, \bibinfo{pages}{593--607}.
\newblock


\bibitem[\protect\citeauthoryear{Schlichtkrull, Kipf, Bloem, Van Den~Berg,
  Titov, and Welling}{Schlichtkrull et~al\mbox{.}}{2018b}]%
        {rgcn}
\bibfield{author}{\bibinfo{person}{Michael Schlichtkrull},
  \bibinfo{person}{Thomas~N Kipf}, \bibinfo{person}{Peter Bloem},
  \bibinfo{person}{Rianne Van Den~Berg}, \bibinfo{person}{Ivan Titov}, {and}
  \bibinfo{person}{Max Welling}.} \bibinfo{year}{2018}\natexlab{b}.
\newblock \showarticletitle{Modeling relational data with graph convolutional
  networks}. In \bibinfo{booktitle}{\emph{European semantic web conference}}.
  Springer, \bibinfo{pages}{593--607}.
\newblock


\bibitem[\protect\citeauthoryear{Shi, Hu, Zhao, and Philip}{Shi
  et~al\mbox{.}}{2018}]%
        {related:heRec}
\bibfield{author}{\bibinfo{person}{Chuan Shi}, \bibinfo{person}{Binbin Hu},
  \bibinfo{person}{Wayne~Xin Zhao}, {and} \bibinfo{person}{S~Yu Philip}.}
  \bibinfo{year}{2018}\natexlab{}.
\newblock \showarticletitle{Heterogeneous information network embedding for
  recommendation}.
\newblock \bibinfo{journal}{\emph{IEEE Transactions on Knowledge and Data
  Engineering}} \bibinfo{volume}{31}, \bibinfo{number}{2}
  (\bibinfo{year}{2018}), \bibinfo{pages}{357--370}.
\newblock


\bibitem[\protect\citeauthoryear{Shi, Li, Zhang, Sun, and Philip}{Shi
  et~al\mbox{.}}{2016}]%
        {heterogeneous-embb-survey}
\bibfield{author}{\bibinfo{person}{Chuan Shi}, \bibinfo{person}{Yitong Li},
  \bibinfo{person}{Jiawei Zhang}, \bibinfo{person}{Yizhou Sun}, {and}
  \bibinfo{person}{S~Yu Philip}.} \bibinfo{year}{2016}\natexlab{}.
\newblock \showarticletitle{A survey of heterogeneous information network
  analysis}.
\newblock \bibinfo{journal}{\emph{IEEE Transactions on Knowledge and Data
  Engineering}} \bibinfo{volume}{29}, \bibinfo{number}{1}
  (\bibinfo{year}{2016}), \bibinfo{pages}{17--37}.
\newblock


\bibitem[\protect\citeauthoryear{Sun and Han}{Sun and Han}{2013}]%
        {HINDef}
\bibfield{author}{\bibinfo{person}{Yizhou Sun} {and} \bibinfo{person}{Jiawei
  Han}.} \bibinfo{year}{2013}\natexlab{}.
\newblock \showarticletitle{Mining heterogeneous information networks: a
  structural analysis approach}.
\newblock \bibinfo{journal}{\emph{Acm Sigkdd Explorations Newsletter}}
  \bibinfo{volume}{14}, \bibinfo{number}{2} (\bibinfo{year}{2013}),
  \bibinfo{pages}{20--28}.
\newblock


\bibitem[\protect\citeauthoryear{Sun, Han, Yan, Yu, and Wu}{Sun
  et~al\mbox{.}}{2011}]%
        {metapathDef}
\bibfield{author}{\bibinfo{person}{Yizhou Sun}, \bibinfo{person}{Jiawei Han},
  \bibinfo{person}{Xifeng Yan}, \bibinfo{person}{Philip~S Yu}, {and}
  \bibinfo{person}{Tianyi Wu}.} \bibinfo{year}{2011}\natexlab{}.
\newblock \showarticletitle{Pathsim: Meta path-based top-k similarity search in
  heterogeneous information networks}.
\newblock \bibinfo{journal}{\emph{Proceedings of the VLDB Endowment}}
  \bibinfo{volume}{4}, \bibinfo{number}{11} (\bibinfo{year}{2011}),
  \bibinfo{pages}{992--1003}.
\newblock


\bibitem[\protect\citeauthoryear{Tang, Qu, Wang, Zhang, Yan, and Mei}{Tang
  et~al\mbox{.}}{2015}]%
        {related:line}
\bibfield{author}{\bibinfo{person}{Jian Tang}, \bibinfo{person}{Meng Qu},
  \bibinfo{person}{Mingzhe Wang}, \bibinfo{person}{Ming Zhang},
  \bibinfo{person}{Jun Yan}, {and} \bibinfo{person}{Qiaozhu Mei}.}
  \bibinfo{year}{2015}\natexlab{}.
\newblock \showarticletitle{Line: Large-scale information network embedding}.
  In \bibinfo{booktitle}{\emph{Proceedings of the 24th international conference
  on world wide web}}. \bibinfo{pages}{1067--1077}.
\newblock


\bibitem[\protect\citeauthoryear{Tian, Gao, Cui, Chen, and Liu}{Tian
  et~al\mbox{.}}{2014}]%
        {related-homo-cls1}
\bibfield{author}{\bibinfo{person}{Fei Tian}, \bibinfo{person}{Bin Gao},
  \bibinfo{person}{Qing Cui}, \bibinfo{person}{Enhong Chen}, {and}
  \bibinfo{person}{Tie-Yan Liu}.} \bibinfo{year}{2014}\natexlab{}.
\newblock \showarticletitle{Learning deep representations for graph
  clustering}. In \bibinfo{booktitle}{\emph{Proceedings of the AAAI Conference
  on Artificial Intelligence}}, Vol.~\bibinfo{volume}{28}.
\newblock


\bibitem[\protect\citeauthoryear{Tsitsulin, Palowitch, Perozzi, and
  M{\"u}ller}{Tsitsulin et~al\mbox{.}}{2020}]%
        {related-homo-cls2}
\bibfield{author}{\bibinfo{person}{Anton Tsitsulin}, \bibinfo{person}{John
  Palowitch}, \bibinfo{person}{Bryan Perozzi}, {and} \bibinfo{person}{Emmanuel
  M{\"u}ller}.} \bibinfo{year}{2020}\natexlab{}.
\newblock \showarticletitle{Graph clustering with graph neural networks}.
\newblock \bibinfo{journal}{\emph{arXiv preprint arXiv:2006.16904}}
  (\bibinfo{year}{2020}).
\newblock


\bibitem[\protect\citeauthoryear{Tu, Zeng, Wang, Zhang, Liu, Sun, Zhang, and
  Lin}{Tu et~al\mbox{.}}{2018b}]%
        {cnrl}
\bibfield{author}{\bibinfo{person}{Cunchao Tu}, \bibinfo{person}{Xiangkai
  Zeng}, \bibinfo{person}{Hao Wang}, \bibinfo{person}{Zhengyan Zhang},
  \bibinfo{person}{Zhiyuan Liu}, \bibinfo{person}{Maosong Sun},
  \bibinfo{person}{Bo Zhang}, {and} \bibinfo{person}{Leyu Lin}.}
  \bibinfo{year}{2018}\natexlab{b}.
\newblock \showarticletitle{A unified framework for community detection and
  network representation learning}.
\newblock \bibinfo{journal}{\emph{IEEE Transactions on Knowledge and Data
  Engineering}} \bibinfo{volume}{31}, \bibinfo{number}{6}
  (\bibinfo{year}{2018}), \bibinfo{pages}{1051--1065}.
\newblock


\bibitem[\protect\citeauthoryear{Tu, Cui, Wang, Wang, and Zhu}{Tu
  et~al\mbox{.}}{2018a}]%
        {related:dhne}
\bibfield{author}{\bibinfo{person}{Ke Tu}, \bibinfo{person}{Peng Cui},
  \bibinfo{person}{Xiao Wang}, \bibinfo{person}{Fei Wang}, {and}
  \bibinfo{person}{Wenwu Zhu}.} \bibinfo{year}{2018}\natexlab{a}.
\newblock \showarticletitle{Structural deep embedding for hyper-networks}. In
  \bibinfo{booktitle}{\emph{Proceedings of the AAAI Conference on Artificial
  Intelligence}}, Vol.~\bibinfo{volume}{32}.
\newblock


\bibitem[\protect\citeauthoryear{Veli{\v{c}}kovi{\'c}, Cucurull, Casanova,
  Romero, Lio, and Bengio}{Veli{\v{c}}kovi{\'c} et~al\mbox{.}}{2017}]%
        {gat}
\bibfield{author}{\bibinfo{person}{Petar Veli{\v{c}}kovi{\'c}},
  \bibinfo{person}{Guillem Cucurull}, \bibinfo{person}{Arantxa Casanova},
  \bibinfo{person}{Adriana Romero}, \bibinfo{person}{Pietro Lio}, {and}
  \bibinfo{person}{Yoshua Bengio}.} \bibinfo{year}{2017}\natexlab{}.
\newblock \showarticletitle{Graph attention networks}.
\newblock \bibinfo{journal}{\emph{arXiv preprint arXiv:1710.10903}}
  (\bibinfo{year}{2017}).
\newblock


\bibitem[\protect\citeauthoryear{Velickovic, Fedus, Hamilton, Li{\`o}, Bengio,
  and Hjelm}{Velickovic et~al\mbox{.}}{2019}]%
        {dgi}
\bibfield{author}{\bibinfo{person}{Petar Velickovic}, \bibinfo{person}{William
  Fedus}, \bibinfo{person}{William~L Hamilton}, \bibinfo{person}{Pietro
  Li{\`o}}, \bibinfo{person}{Yoshua Bengio}, {and} \bibinfo{person}{R~Devon
  Hjelm}.} \bibinfo{year}{2019}\natexlab{}.
\newblock \showarticletitle{Deep graph infomax}.
\newblock  (\bibinfo{year}{2019}).
\newblock


\bibitem[\protect\citeauthoryear{Wang, Zheng, Ye, Gan, Li, Song, Zhou, Ma, Yu,
  Gai, Xiao, He, Karypis, Li, and Zhang}{Wang et~al\mbox{.}}{2019b}]%
        {dgl}
\bibfield{author}{\bibinfo{person}{Minjie Wang}, \bibinfo{person}{Da Zheng},
  \bibinfo{person}{Zihao Ye}, \bibinfo{person}{Quan Gan},
  \bibinfo{person}{Mufei Li}, \bibinfo{person}{Xiang Song},
  \bibinfo{person}{Jinjing Zhou}, \bibinfo{person}{Chao Ma},
  \bibinfo{person}{Lingfan Yu}, \bibinfo{person}{Yu Gai},
  \bibinfo{person}{Tianjun Xiao}, \bibinfo{person}{Tong He},
  \bibinfo{person}{George Karypis}, \bibinfo{person}{Jinyang Li}, {and}
  \bibinfo{person}{Zheng Zhang}.} \bibinfo{year}{2019}\natexlab{b}.
\newblock \showarticletitle{Deep Graph Library: A Graph-Centric,
  Highly-Performant Package for Graph Neural Networks}.
\newblock \bibinfo{journal}{\emph{arXiv preprint arXiv:1909.01315}}
  (\bibinfo{year}{2019}).
\newblock


\bibitem[\protect\citeauthoryear{Wang, Ji, Shi, Wang, Ye, Cui, and Yu}{Wang
  et~al\mbox{.}}{2019a}]%
        {heteroGat}
\bibfield{author}{\bibinfo{person}{Xiao Wang}, \bibinfo{person}{Houye Ji},
  \bibinfo{person}{Chuan Shi}, \bibinfo{person}{Bai Wang},
  \bibinfo{person}{Yanfang Ye}, \bibinfo{person}{Peng Cui}, {and}
  \bibinfo{person}{Philip~S Yu}.} \bibinfo{year}{2019}\natexlab{a}.
\newblock \showarticletitle{Heterogeneous graph attention network}. In
  \bibinfo{booktitle}{\emph{The World Wide Web Conference}}.
  \bibinfo{pages}{2022--2032}.
\newblock


\bibitem[\protect\citeauthoryear{Wu, Lian, Xu, Wu, and Chen}{Wu
  et~al\mbox{.}}{2020}]%
        {applications-social5}
\bibfield{author}{\bibinfo{person}{Yongji Wu}, \bibinfo{person}{Defu Lian},
  \bibinfo{person}{Yiheng Xu}, \bibinfo{person}{Le Wu}, {and}
  \bibinfo{person}{Enhong Chen}.} \bibinfo{year}{2020}\natexlab{}.
\newblock \showarticletitle{Graph convolutional networks with markov random
  field reasoning for social spammer detection}. In
  \bibinfo{booktitle}{\emph{Proceedings of the AAAI Conference on Artificial
  Intelligence}}, Vol.~\bibinfo{volume}{34}. \bibinfo{pages}{1054--1061}.
\newblock


\bibitem[\protect\citeauthoryear{Wu, Xu, Chen, and Wang}{Wu
  et~al\mbox{.}}{2005}]%
        {applications-others1}
\bibfield{author}{\bibinfo{person}{Zhi-Xi Wu}, \bibinfo{person}{Xin-Jian Xu},
  \bibinfo{person}{Yong Chen}, {and} \bibinfo{person}{Ying-Hai Wang}.}
  \bibinfo{year}{2005}\natexlab{}.
\newblock \showarticletitle{Spatial prisoner’s dilemma game with volunteering
  in Newman-Watts small-world networks}.
\newblock \bibinfo{journal}{\emph{Physical Review E}} \bibinfo{volume}{71},
  \bibinfo{number}{3} (\bibinfo{year}{2005}), \bibinfo{pages}{037103}.
\newblock


\bibitem[\protect\citeauthoryear{Xie, Girshick, and Farhadi}{Xie
  et~al\mbox{.}}{2016}]%
        {euclidean-joint1}
\bibfield{author}{\bibinfo{person}{Junyuan Xie}, \bibinfo{person}{Ross
  Girshick}, {and} \bibinfo{person}{Ali Farhadi}.}
  \bibinfo{year}{2016}\natexlab{}.
\newblock \showarticletitle{Unsupervised deep embedding for clustering
  analysis}. In \bibinfo{booktitle}{\emph{International conference on machine
  learning}}. PMLR, \bibinfo{pages}{478--487}.
\newblock


\bibitem[\protect\citeauthoryear{Yang, Fu, Sidiropoulos, and Hong}{Yang
  et~al\mbox{.}}{2017}]%
        {euclidean-joint2}
\bibfield{author}{\bibinfo{person}{Bo Yang}, \bibinfo{person}{Xiao Fu},
  \bibinfo{person}{Nicholas~D Sidiropoulos}, {and} \bibinfo{person}{Mingyi
  Hong}.} \bibinfo{year}{2017}\natexlab{}.
\newblock \showarticletitle{Towards k-means-friendly spaces: Simultaneous deep
  learning and clustering}. In \bibinfo{booktitle}{\emph{international
  conference on machine learning}}. PMLR, \bibinfo{pages}{3861--3870}.
\newblock


\bibitem[\protect\citeauthoryear{Ying, He, Chen, Eksombatchai, Hamilton, and
  Leskovec}{Ying et~al\mbox{.}}{2018}]%
        {applications-social3}
\bibfield{author}{\bibinfo{person}{Rex Ying}, \bibinfo{person}{Ruining He},
  \bibinfo{person}{Kaifeng Chen}, \bibinfo{person}{Pong Eksombatchai},
  \bibinfo{person}{William~L Hamilton}, {and} \bibinfo{person}{Jure Leskovec}.}
  \bibinfo{year}{2018}\natexlab{}.
\newblock \showarticletitle{Graph convolutional neural networks for web-scale
  recommender systems}. In \bibinfo{booktitle}{\emph{Proceedings of the 24th
  ACM SIGKDD International Conference on Knowledge Discovery \& Data Mining}}.
  \bibinfo{pages}{974--983}.
\newblock


\bibitem[\protect\citeauthoryear{Zhao, Wang, Shi, Liu, and Ye}{Zhao
  et~al\mbox{.}}{2020}]%
        {nshe}
\bibfield{author}{\bibinfo{person}{Jianan Zhao}, \bibinfo{person}{Xiao Wang},
  \bibinfo{person}{Chuan Shi}, \bibinfo{person}{Zekuan Liu}, {and}
  \bibinfo{person}{Yanfang Ye}.} \bibinfo{year}{2020}\natexlab{}.
\newblock \showarticletitle{Network Schema Preserving Heterogeneous Information
  Network Embedding}. IJCAI.
\newblock


\bibitem[\protect\citeauthoryear{Zhou, Cui, Hu, Zhang, Yang, Liu, Wang, Li, and
  Sun}{Zhou et~al\mbox{.}}{2020}]%
        {applications-survey1}
\bibfield{author}{\bibinfo{person}{Jie Zhou}, \bibinfo{person}{Ganqu Cui},
  \bibinfo{person}{Shengding Hu}, \bibinfo{person}{Zhengyan Zhang},
  \bibinfo{person}{Cheng Yang}, \bibinfo{person}{Zhiyuan Liu},
  \bibinfo{person}{Lifeng Wang}, \bibinfo{person}{Changcheng Li}, {and}
  \bibinfo{person}{Maosong Sun}.} \bibinfo{year}{2020}\natexlab{}.
\newblock \showarticletitle{Graph neural networks: A review of methods and
  applications}.
\newblock \bibinfo{journal}{\emph{AI Open}}  \bibinfo{volume}{1}
  (\bibinfo{year}{2020}), \bibinfo{pages}{57--81}.
\newblock


\bibitem[\protect\citeauthoryear{Zhou, Ma, Lyu, and King}{Zhou
  et~al\mbox{.}}{2010}]%
        {applications-social1}
\bibfield{author}{\bibinfo{person}{Tom Zhou}, \bibinfo{person}{Hao Ma},
  \bibinfo{person}{Michael Lyu}, {and} \bibinfo{person}{Irwin King}.}
  \bibinfo{year}{2010}\natexlab{}.
\newblock \showarticletitle{Userrec: A user recommendation framework in social
  tagging systems}. In \bibinfo{booktitle}{\emph{Proceedings of the AAAI
  Conference on Artificial Intelligence}}, Vol.~\bibinfo{volume}{24}.
\newblock


\bibitem[\protect\citeauthoryear{Zhu, Byrd, Lu, and Nocedal}{Zhu
  et~al\mbox{.}}{1997}]%
        {lbfgs}
\bibfield{author}{\bibinfo{person}{Ciyou Zhu}, \bibinfo{person}{Richard~H.
  Byrd}, \bibinfo{person}{Peihuang Lu}, {and} \bibinfo{person}{Jorge Nocedal}.}
  \bibinfo{year}{1997}\natexlab{}.
\newblock \showarticletitle{Algorithm 778: L-BFGS-B: Fortran Subroutines for
  Large-Scale Bound-Constrained Optimization}.
\newblock \bibinfo{journal}{\emph{ACM Trans. Math. Softw.}}
  (\bibinfo{year}{1997}).
\newblock


\end{thebibliography}

\clearpage
\appendix
\section*{\LARGE{SUPPLEMENTARY MATERIAL}}
Throughout the following sections, we make use of the notation and the references from the paper.

\section{Derivation of ELBO Bound} \label{sec:derivationOfELBO}

\begin{align}
	&\text{log}(p_{\theta}(\mA)) = \text{log}\bigg(\int \sum \limits_{\vc} p_{\theta}(\mZ, \vc, \mA) d\mZ\bigg) \\
	&= \text{log}\bigg(\int \sum \limits_{\vc} p(\mZ) \ p_{\theta}(\vc | \mZ) \ p_{\theta}(\mA | \vc, \mZ)d\mZ\bigg) \\
	&= \text{log}\bigg( \mathbb{E}_{(\mZ, \vc) \sim q_{\phi}(\mZ, \vc | \gI)} \bigg\{\frac{p(\mZ) \ p_{\theta}(\vc | \mZ) \ p_{\theta}(\mA | \vc, \mZ)}{q_{\phi}(\mZ | \gI) q_{\phi}(\vc | \mZ, \gI)}\bigg\} \bigg) \\
	&\geq \mathbb{E}_{(\mZ, \vc) \sim q_{\phi}(\mZ, \vc | \gI)} \bigg\{ \text{log}\bigg( \frac{p(\mZ) \ p_{\theta}(\vc | \mZ) \ p_{\theta}(\mA | \vc, \mZ)}{q_{\phi}(\mZ | \gI) q_{\phi}(\vc | \mZ, \gI)}\bigg)\bigg\} \label{eq:obj:jensen-applied} \\
    &= \mathbb{E}_{(\mZ, \vc) \sim q_{\phi}(\mZ, \vc | \gI)} \bigg\{ \text{log}\bigg(\frac{p(\mZ)}{q_{\phi}(\mZ | \gI)}\bigg) \nonumber \\
    &+ \text{log}\bigg(\frac{p_{\theta}(\vc | \mZ)}{q_{\phi}(\vc | \mZ, \gI)}\bigg) + \text{log}\bigg(p_{\theta}(\mA | \vc, \mZ)\bigg) \bigg\} \\
	&= \mathbb{E}_{\mZ \sim q_{\phi}(\mZ | \gI)} \bigg\{ \text{log}\bigg(\frac{p(\mZ)}{q_{\phi}(\mZ | \gI)}\bigg) \bigg\} \nonumber \\
	&+ \mathbb{E}_{(\mZ, \vc) \sim q_{\phi}(\mZ, \vc | \gI)} \bigg\{ \text{log}\bigg(\frac{p_{\theta}(\vc | \mZ)}{q_{\phi}(\vc | \mZ, \gI)}\bigg) \bigg\} \nonumber \\
	&+ \mathbb{E}_{(\mZ, \vc) \sim q_{\phi}(\mZ, \vc | \gI)} \bigg\{ \text{log}\bigg(p_{\theta}(\mA | \vc, \mZ)\bigg) \bigg\}. \label{eq:obj:complete}
\end{align}

Here (\ref{eq:obj:jensen-applied}) follows from Jensen's Inequality.
The first term of (\ref{eq:obj:complete}) is given by:
\begin{align}
    &\mathbb{E}_{\mZ \sim q_{\phi}(\mZ | \gI)} \bigg\{ \text{log}\bigg(\frac{p(\mZ)}{q_{\phi}(\mZ | \gI)}\bigg) \bigg\} \nonumber \\ 
    &= \sum \limits_{i=1}^{N} \mathbb{E}_{\vz_{i} \sim q_{\phi}(\vz_{i} | \gI)} \bigg\{ \text{log}\bigg(\frac{p(\vz_{i})}{q_{\phi}(\vz_{i} | \gI)}\bigg) \bigg\} \\
    &= - \sum \limits_{i=1}^{N} D_{KL} (q_{\phi}(\vz_{i} | \gI) \ || \ p(\vz_{i})) \label{eq:kl_z} \\
    &= -L_{\mathrm{enc}}. 
\end{align}
The second term of \eqref{eq:obj:complete} can be derived as:
\begin{align}
    &\mathbb{E}_{(\mZ, \vc) \sim q_{\phi}(\mZ, \vc | \gI)} \bigg\{ \text{log}\bigg(\frac{p_{\theta}(\vc | \mZ)}{q_{\phi}(\vc | \mZ, \gI)}\bigg) \bigg\} \nonumber \\
    &= \sum \limits_{i=1}^{N}  \mathbb{E}_{(\vz_{i}, c_{i}) \sim q_{\phi}(\vz_{i}, c_{i} | \gI)} \bigg\{\text{log}\bigg(\frac{p_{\theta}(c_{i} | \vz_{i})}{q_{\phi}(c_{i} | \vz_{i}, \gI)}\bigg) \bigg\} \label{eq:kl_c_iner_1} \\
    &\approx \sum \limits_{i=1}^{N} \frac{1}{R} \sum \limits_{r = 1}^{R} \mathbb{E}_{c_{i} \sim q_{\phi}(c_{i} | \vz_{i}^{(r)}, \gI) } \bigg\{\text{log}\bigg(\frac{p_{\theta}(c_{i} | \vz_{i}^{(r)})}{q_{\phi}(c_{i} | \vz_{i}^{(r)}, \gI)}\bigg) \bigg\} \label{eq:kl_c_inter_2} \\
    &= - \sum \limits_{i=1}^{N} \frac{1}{R} \sum \limits_{r = 1}^{R} D_{KL} (q_{\phi}(c_{i} | \vz_{i}^{(r)}, \gI) \ ||\ p_{\theta}(c_{i} | \vz_{i}^{(r)})) \label{eq:kl_c} \\
    &= -L_c.
\end{align}
Here (\ref{eq:kl_c_inter_2}) follows from \eqref{eq:kl_c_iner_1} by replacing the expectation over $\vz_{i}$ with sample mean by generating $R$ samples $\vz_{i}^{(r)}$ from distribution $q(\vz_{i} | \gI)$.
Assuming $\erva_{ij} \in [0, 1] \ \forall i$, the third term of \eqref{eq:obj:complete} is the negative of binary cross entropy (BCE) between observed and predicted edges. 

\begin{align}
    &\mathbb{E}_{(\mZ, \vc) \sim q_{\phi}(\mZ, \vc | \gI)} \bigg\{ \text{log}\bigg(p_{\theta}(\mA | \vc, \mZ)\bigg) \bigg\} \nonumber \\
    &\approx \sum \limits_{(i, j) \in \gE} \mathbb{E}_{q_{\phi}(\vz_{i}, \vz_{j}, c_{i}, c_{j} | \gI)}\bigg\{\text{log}\bigg(p_{\theta}(a_{ij} | c_{i}, c_{j}, \vz_{i}, \vz_{j})\bigg)\bigg\} \\
    & = -L_{\mathrm{recon}}. \label{eq:bce}
\end{align}

Hence, by substituting \eqref{eq:kl_z} and \eqref{eq:kl_c} in \eqref{eq:obj:complete}, we get the ELBO bound as:

\begin{align}
    \mathcal{L}_{ELBO}& \approx -\sum \limits_{i=1}^{N} D_{KL} (q_{\phi}(\vz_{i} | \gI) \ || \ p(\vz_{i})) \nonumber \\
    -& \sum \limits_{i=1}^{N} \frac{1}{R} \sum \limits_{r = 1}^{R} D_{KL} (q_{\phi}(c_{i} | \vz_{i}^{(r)}, \gI) \ ||\ p_{\theta}(c_{i} | \vz_{i}^{(r)})) \nonumber \\
    +& \sum \limits_{(i, j) \in \gE} \mathbb{E}_{q_{\phi}(\vz_{i}, \vz_{j}, c_{i}, c_{j} | \gI)}\bigg\{\text{log}\bigg( p_{\theta}(a_{ij} | c_{i}, c_{j}, \vz_{i}, \vz_{j})\bigg)\bigg\}
\end{align}

\section{\ours with Different Encoders}
In this section we evaluate the performance of \ours for five different encoders.
For GCN\cite{exp:split-2} and RGCN\cite{rgcn}, we compare both variational and non-variational counterparts.
For variational encoder, we learn both mean and variance, followed by reparameterization trick as given in \secref{sec:LencConsiderations}.
For non-variational encoders, we ignore $L_{\mathrm{enc}}$ as we directly learn $\mZ$ from the input $\gI$.
In addition to GCN and RGCN encoders, we also give the results for a simple linear encoder consisting of two matrices to learn the parameters of $q_{\phi}(\vz_i|\gI)$.
For classification, we only report F1-Micro score as F1-Macro follows the same trend.

\begin{figure*}
    \centering
    \begin{subfigure}[t]{0.45\linewidth}
        \includegraphics[width=\linewidth]{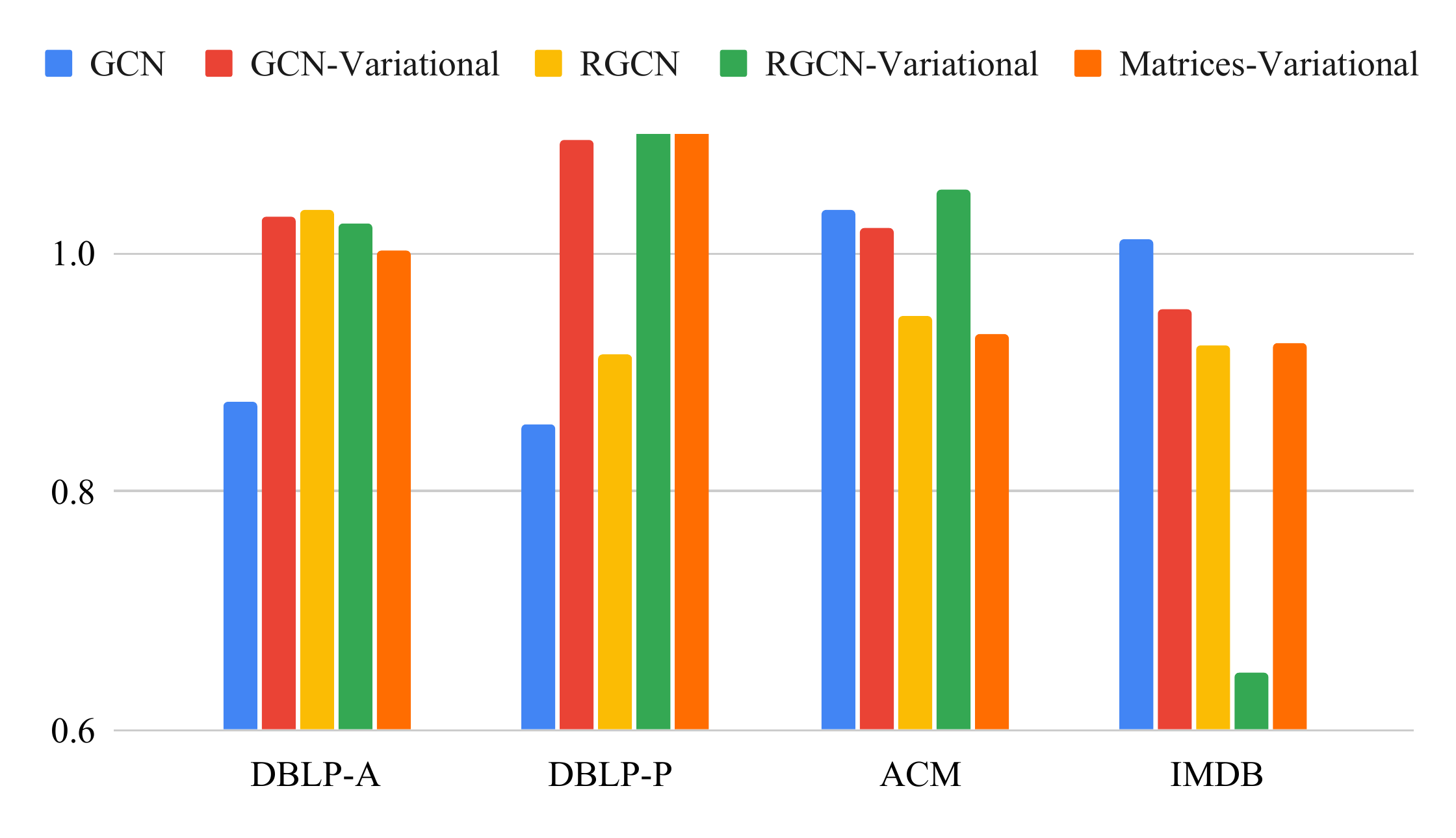}
        \caption{With $L_{\mathrm{cont}}$.}
    \end{subfigure}
    \begin{subfigure}[t]{0.45\linewidth}
        \includegraphics[width=\linewidth]{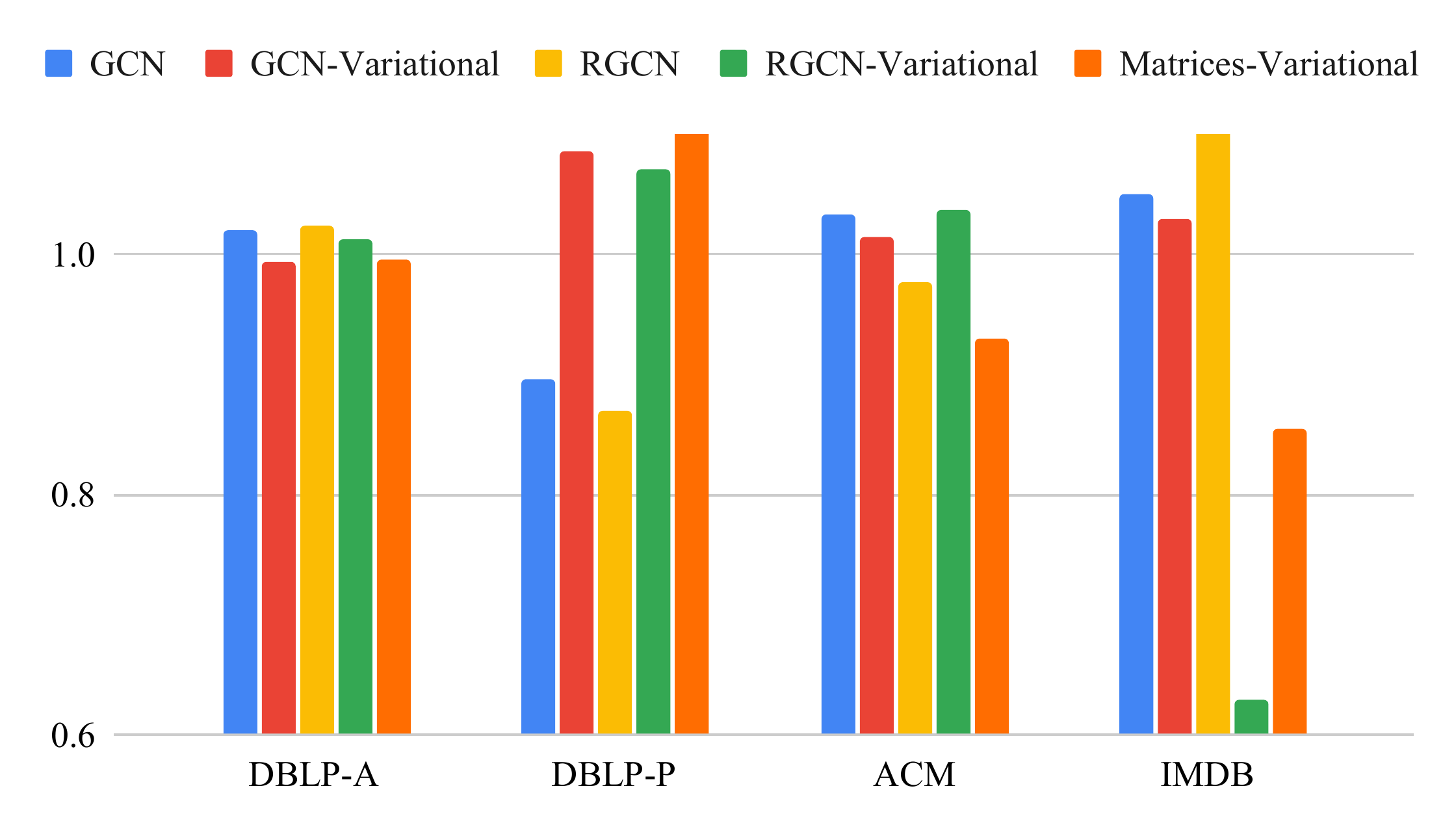}
        \caption{Without $L_{\mathrm{cont}}$.}
    \end{subfigure}
    \caption{Clustering performance (NMI) of \ours for different types of encoders.}
    \label{fig:clusteringDifferentEncoders}
\end{figure*}

\begin{figure*}
    \centering
    \begin{subfigure}[t]{0.45\linewidth}
        \includegraphics[width=\linewidth]{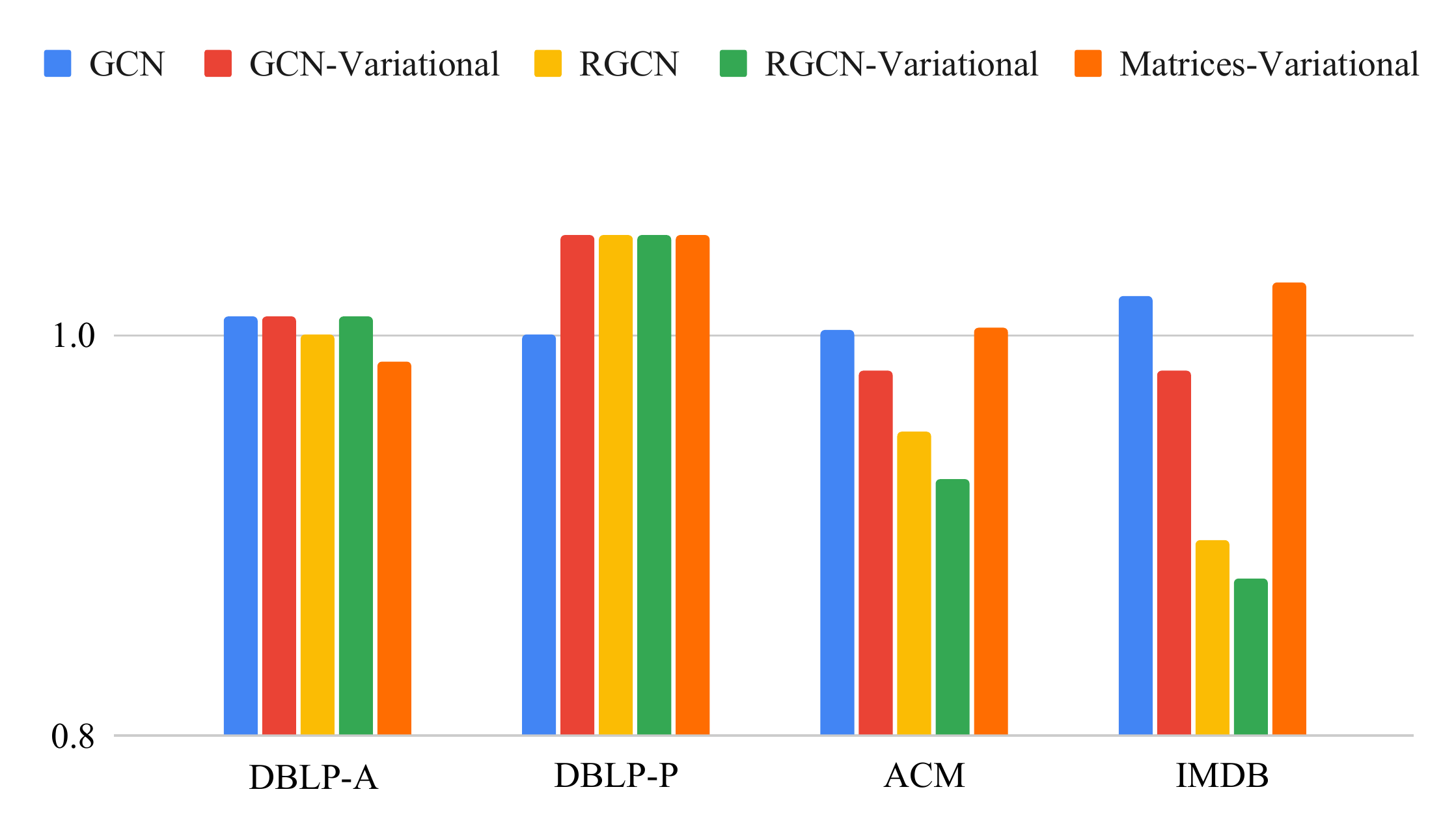}
        \caption{With $L_{\mathrm{cont}}$.}
        \end{subfigure}
    \begin{subfigure}[t]{0.45\linewidth}
        \includegraphics[width=\linewidth]{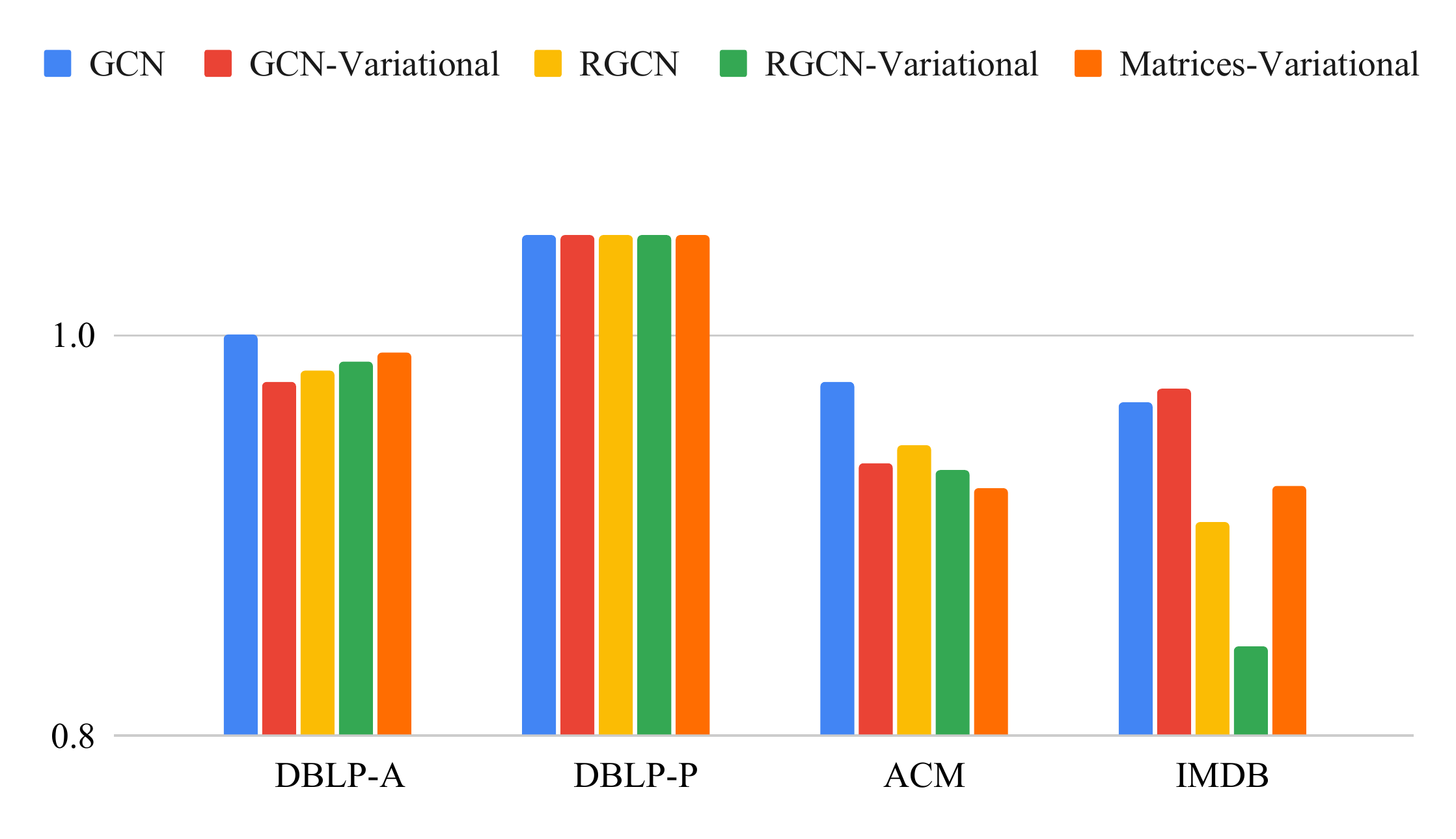}
        \caption{Without $L_{\mathrm{cont}}$.}
        \end{subfigure}
    \caption{Classification performance (F1-Micro) of \ours for different types of encoders.}
    \label{fig:classificationDifferentEncoders}
\end{figure*}

The relative clustering and classification performances achieved using these encoders are illustrated in \figref{fig:clusteringDifferentEncoders} and \figref{fig:classificationDifferentEncoders} respectively.
The results in these bar plots are normalized by the second best values in \tabref{tab:resClustering} and \tabref{tab:resClassification} for better visualization.
In \figref{fig:clusteringDifferentEncoders} the results are usually better with $L_{\mathrm{cont}}$, although the performance difference is rather small.
One exception is the IMDB dataset where the results are better without $L_{\mathrm{cont}}$ as stated in \secref{sec:experimentsNodeClustering}.
Overall, \ours performs better than its competitors with at least three out of five encoders in all four cases.
The effect of $L_{\mathrm{cont}}$ gets more highlighted for the classification task in \figref{fig:classificationDifferentEncoders}.
For instance, without $L_{\mathrm{cont}}$, \ours fails to beat the second-best F1-Micro scores for ACM and IMDB, irrespective of the chosen encoder architecture.
In addition, it is worth noticing that even a simple matrices-based variational encoder yields reasonable performance for downstream classification, which hints on the stability of the architecture of \ours for downstream node classification even with a simple encoder.


\end{document}